\pdfoutput=1

\documentclass[11pt]{article}

\usepackage{emnlp2021}

\usepackage{times}
\usepackage{latexsym}

\usepackage[T1]{fontenc}

\usepackage[utf8]{inputenc}

\usepackage{microtype}
\usepackage{microtype}
\usepackage{graphicx}
\usepackage{subfigure}
\usepackage{booktabs} 
\usepackage{footnote}

\usepackage{hyperref}

\usepackage{amsmath}
\usepackage{amssymb}
\usepackage{enumitem}
\usepackage{graphicx}
\usepackage{xspace}
\usepackage{array,multirow}
\usepackage{xargs} 
\usepackage{url}
\usepackage{ragged2e}
\usepackage[normalem]{ulem}

\newcommand{\roberta}{\textsc{Roberta}\xspace}
\newcommand{\bigb}{\textsc{BigBird}\xspace}
\newcommand{\spql}{\textsc{Sparql}\xspace}
\newcommand{\hitl}{\textsc{H-I-T-L}\xspace}
\newcommand{\webqsp}{\textsc{WebQuestionsSP}\xspace}
\newcommand{\cwq}{\textsc{ComplexWebQuestions}\xspace}

\newcommand{\alg}{\textsc{Cbr-kbqa}\xspace}

\newcommand{\ignore}[1]{}

\title{Case-Based Reasoning for Natural Language Queries \\over Knowledge Bases}

\author{Rajarshi Das$^{1}$, Manzil Zaheer$^{3}$, Dung Thai$^{1}$, Ameya Godbole$^{1}$, Ethan Perez$^{2}$, \\\textbf{Jay-Yoon Lee$^{1}$, Lizhen Tan$^{4}$, Lazaros Polymenakos$^{5}$\thanks{$^*$Now director of AI at EY CESA, Greece. Work done while at Amazon}, Andrew McCallum$^{1}$}\\
$^{1}$University of Massachusetts, Amherst, $^{2}$New York University\\
$^{3}$Google Research, $^{4}$Amazon, $^{5}$EY, Greece\\
\texttt{\{rajarshi}, \texttt{dthai}, \texttt{agodbole}, \texttt{jaylee}, \texttt{mccallum\}}\texttt{@cs.umass.edu}
}

\begin{document}
\maketitle
\begin{abstract}

It is often challenging to solve a complex problem from scratch, but much easier if we can access other similar problems with their solutions --- a paradigm known as case-based reasoning (CBR).
We propose a neuro-symbolic CBR approach (\alg) for question answering over large knowledge bases. \alg consists of a nonparametric memory that stores cases (question and logical forms) and a parametric model that can generate a logical form for a new question by retrieving cases that are relevant to it. On several KBQA datasets that contain complex questions, \alg achieves competitive performance. For example, on the \cwq dataset, \alg outperforms the current state of the art by 11\% on accuracy. Furthermore, we show that \alg is capable of using new cases \emph{without} any further training: by incorporating a few human-labeled examples in the case memory, \alg is able to successfully generate logical forms containing unseen KB entities as well as relations.

\end{abstract}

\section{Introduction}
\label{sec:intro}
Humans often solve a new problem by recollecting and adapting the solution to multiple related problems that they have encountered in the past \cite{ross1984remindings,lancaster1987problem,schmidt1990cognitive}. In classical artificial intelligence (AI), case-based reasoning (CBR) pioneered by \citet{schank1982dynamic}, tries to incorporate such model of reasoning in AI systems \cite{kolodner1983maintaining,rissland1983examples,leake1996cbr}. A sketch of a CBR system \cite{aamodt1994case} comprises of --- (i) a retrieval module, in which `cases' that are similar to the given problem are retrieved, (ii) a reuse module, where the solutions of the retrieved cases are re-used to synthesize a new solution. Often, the new solution does not work and needs more revision, which is handled by a (iii) revise module.

In its early days, the components of CBR were implemented with symbolic systems, which had their limitations. For example, finding similar cases and synthesizing new solutions from them is a challenging task for a CBR system implemented with symbolic components. However, with the recent advancements in representation learning \cite{lecun2015deep}, the performance of ML systems have improved substantially on a range of practical tasks.  

\begin{figure*}
    \centering
    \includegraphics[width=2\columnwidth]{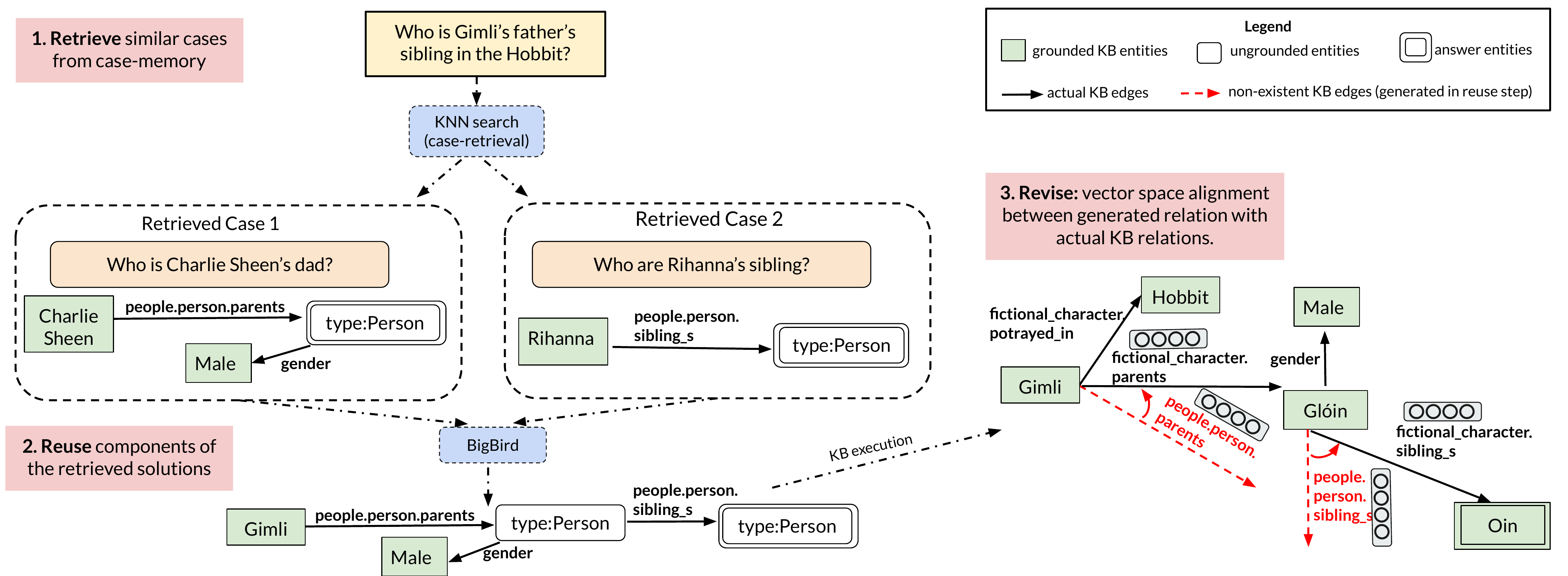}
    \vspace{-1mm}
    \caption{\alg derives the logical form (LF) for a new query from the LFs of other retrieved queries from the case-memory. However, the derived LF might not execute because of missing edges in the KB. The revise step aligns any such missing edges (relations) with existing semantically-similar edges in the KB.}
    \label{fig:intro2}
\end{figure*}

Given a query, \alg uses a neural retriever to retrieve other similar queries (and their logical forms) from a case memory (e.g. training set). Next, \alg generates a logical form for the given query by learning to reuse various components of the logical forms of the retrieved cases. However, often the generated logical form does not produce the right answer when executed against a knowledge base (KB). This can happen because one or more KB relations needed are never present in the retrieved cases or because KBs are woefully incomplete \cite{min2013distant} (Figure~\ref{fig:intro2}). To alleviate such cases, \alg has an additional revise step that \emph{aligns} the generated relations in the logical form to the query entities' local neighborhood in the KB. To achieve this, we take advantage of pre-trained relation embeddings from KB completion techniques (e.g. Trans-E \cite{bordes2013translating}) that learn the structure of the KB.

It has been shown that neural seq2seq models do not generalize well to novel combinations of previously seen input \cite{lake_baroni,loula2018rearranging}. However, \alg has the ability to reuse relations from \emph{multiple} retrieved cases, even if each case contains only partial logic to answer the query. We show that \alg is effective for questions that need novel combination of KB relations, achieving competitive results on multiple KBQA benchmarks such as  WebQuestionsSP \cite{yih2016value}, ComplexWebQuestions (CWQ) \cite{Talmor2018TheWA} and CompositionalFreebaseQuestions (CFQ) \cite{keysers2020measuring}. For example, on the hidden test-set of the challenging CWQ dataset, \alg outperforms the best system by over 11\% points.

 We further demonstrate that \alg, without the need of any further fine-tuning, also generalizes to queries that need relations which were \emph{never seen} in the training set. This is possible due to \alg's nonparametric approach which allows one to \emph{inject} relevant simple cases during inference, allowing it to reuse new relations from those cases. In a controlled human-in-the-loop experiment, we show that \alg can correctly answer such questions when an expert (e.g. database administrator) injects a few simple cases to the case memory. \alg is able to retrieve those examples from the memory and use the unseen relations to compose new logical forms for the given query.
 
Generalization to unseen KB relations, without any re-training, is out of scope for current neural models. Currently, the popular approach to handle such cases is to re-train or fine-tune the model on new examples. This process is not only time-consuming and laborious but models also suffer from catastrophic forgetting \cite{hinton1987using,kirkpatrick2017overcoming}, making wrong predictions on examples which it previously predicted correctly. We believe that the controllable properties of \alg are essential for QA models to be deployed in real-world settings and hope that our work will inspire further research in this direction.

Recent works such as \textsc{Realm} \cite{guu2020realm} and \textsc{Rag} \cite{lewis2020retrieval} retrieve relevant paragraphs from a nonparametric memory for answering questions. \alg, in contrast, retrieves \emph{similar queries} w.r.t the input query and uses the relational similarity between their logical forms to derive a logical form for the new query. \alg is also similar to the recent retrieve and edit framework \cite{hashimoto2018retrieve} for generating structured output. However, unlike us they condition on only a single retrieved example and hence is unlikely to be able to handle  complex questions that need reuse of partial logic from multiple questions. Moreover, unlike \alg, retrieve and edit does not have a component that can explicitly revise an initially generated output.

The contributions of our paper are as follows --- (a) We present a neural CBR approach for KBQA capable of generating complex logical forms conditioned on similar retrieved questions and their logical forms. (b) Since \alg explicitly learns to reuse cases, we show it is able to generalize to unseen relations at test time, when relevant cases are provided. (c) We also show the efficacy of our revise step of \alg which allows to correct generated output by aligning it to local neighborhood of the query entity. (d) Lastly, we show that \alg significantly outperforms other competitive models on several KBQA benchmarks.

\section{Model}
\label{sec:model}
\begin{figure}
    \centering
    \includegraphics[width=\columnwidth]{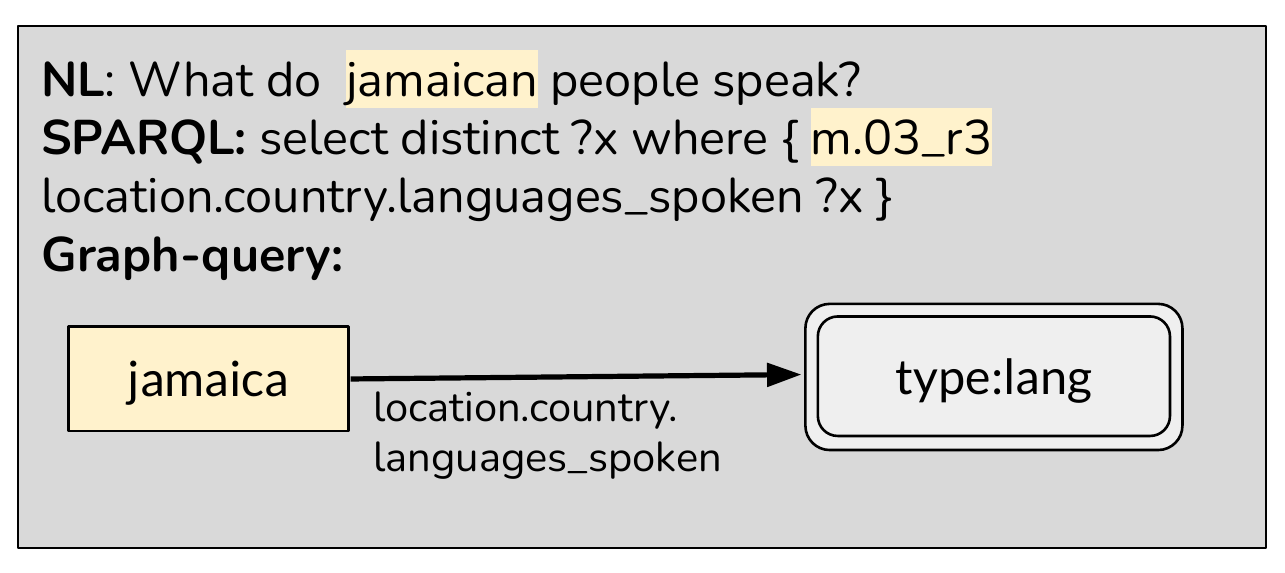}
    \vspace{-5mm}
    \caption{An example of a SPARQL logical form for a simple query and its equivalent graph-query.}
    \label{fig:sparql_example}
\end{figure}

This section describes the implementation of various modules of \alg. In CBR, a case is defined as an abstract representation of a problem along with its solution. In our KBQA setting, a case is a natural language query paired with an executable logical form. The practical importance of KBQA has led to the creation of an array of recent datasets \citep[inter-alia]{zelle1996learning,bordes2015large,su2016generating,yih2016value,zhong2017seq2sql,ngomo20189th,yu2018spider,Talmor2018TheWA}. In these datasets, a question is paired with an executable logical form such as \spql, \textsc{SQL}, S-expression or graph query. All of these forms have equal representational capacity and are interchangeable \cite{su2016generating}. Figure~\ref{fig:sparql_example} shows an example of two equivalent logical forms. For our experiments, we consider \spql programs as our logical form.

\noindent\textbf{Formal definition of task}: let $q$ be a natural language query and let $\mathcal{K}$ be a symbolic KB that needs to be queried to retrieve an answer list $\mathcal{A}$ containing the answer(s) for $q$. We also assume access to a training set $\mathcal{D} = \{(q_1, \ell_1), (q_2, \ell_2), \ldots (q_N, \ell_N)\}$ of queries and their corresponding logical forms where $q_i$, $\ell_i$  represents query and its corresponding logical form, respectively. A logical form is an executable query containing entities, relations and free variables (Figure~\ref{fig:sparql_example}). \alg first retrieves $K$ similar cases $\mathcal{D}_q$  from $\mathcal{D}$ (\S~\ref{sub:retrieval}). It then generates a intermediate logical form $\ell_{\textrm{inter}}$ by learning to reuse components of the logical forms of the retrieved cases (\S~\ref{sub:reuse}). Next, the logical form $\ell_{\textrm{inter}}$ is revised to output the final logical form $\ell$ by aligning to the relations present in the neighborhood subgraph of the query entity to recover from any spurious relations generated in the reuse step (\S~\ref{sub:revise}). Finally, $\ell$ is executed against $\mathcal{K}$ and the list of answer entities are returned. We evaluate our KBQA system by calculating the accuracy of the retrieved answer list w.r.t a held-out set of queries.

\subsection{Retrieve}
\label{sub:retrieval}
The retrieval module computes dense representation of the given query and uses it to retrieve other similar query representation from a training set. Inspired by the recent advances in neural dense passage retrieval \cite{das2019multi,karpukhin2020dense}, we use a \roberta-base encoder to encode each question independently. Also, we want to retrieve questions that have high relational similarity instead of questions which share the same entities (e.g. we prefer to score the query pair (Who is Justin Bieber's brother?, Who is Rihanna's brother?), higher than (Who is Justin Bieber's brother?, Who is Justin Bieber's father?)). To minimize the effect of entities during retrieval, we use a named entity tagger\footnote{\url{https://cloud.google.com/natural-language}} to detect spans of entities and mask them with \textsc{[blank]} symbol with a probability $p_{\textrm{mask}}$, during training. The entity masking strategy has previously been successfully used in learning entity-independent relational representations \cite{soares2019matching}. The similarity score between two queries is given by the inner product between their normalized vector representations (cosine similarity), where each representation, following standard practice \cite{guu2020realm}, is obtained from the encoding of the initial \textsc{[cls]} token of the query. 

\textbf{Fine-tuning question retriever}: In passage retrieval, training data is gathered via distant supervision in which passages containing the answer is marked as a positive example for training. Since in our setup, we need to retrieve similar questions, we use the available logical forms as a source of distant supervision. Specifically, a question pair is weighed by the amount of overlap (w.r.t KB relations) it has in their corresponding logical queries. Following DPR \cite{karpukhin2020dense}, we ensure there is at least one positive example for each query during training and use a weighted negative log-likelihood loss where the weights are computed by the $\textrm{F}_1$ score between the set of relations present in the corresponding logical forms. Concretely, let $(\textrm{q}_1, \textrm{q}_2, \ldots, \textrm{q}_\textrm{B})$ denote all questions in a mini-batch. The loss function is:
\begin{align*}
    L = - \sum_{i, j}w_{i,j}\log\frac{\exp(\mathrm{sim}(\mathbf{q_i, q_j}))}{\sum_{j}\exp(\mathrm{sim}(\mathbf{q_i, q_j}))}
\end{align*}
Here, $\mathbf{q_i} \in \mathbb{R}^{d}$ denotes the vector representation of query 
$\textrm{q}_i$ and $\mathrm{sim}(\mathbf{q_i, q_j}) = \mathbf{q_i^\top} \mathbf{q_j}$. 
$w_{i,j}$ is computed as the F$_{1}$ overlap between relations in the logical pairs of q$_i$ and q$_j$.
We pre-compute and cache the query representations of the training set $\mathcal{D}$. For query $q$, we return the top-$k$ similar queries in $\mathcal{D}$ w.r.t $q$ and pass it to the reuse module.

\subsection{Reuse}
\label{sub:reuse}
The reuse step generates an intermediate logical form from the $k$ cases that are fed to it as input from the retriever module. Pre-trained encoder-decoder transformer models such as \textsc{Bart} \cite{lewis-etal-2020-bart} and T5 \cite{t5} have enjoyed dramatic success on semantic parsing \cite{lin2018nl2bash,hwang2019comprehensive,shaw2020compositional,suhr2020exploring}. We take a similar approach in generating an intermediate logical form conditioned on the retrieved cases. However, one of the core limitation of transformer-based models is its quadratic dependency (in terms of memory), because of full-attention, which severely limits the sequence length it can operate on. For example, \textsc{Bart} and T5 only supports sequence length of 512 tokens in its encoder. Recall that for us, a case is a query from the train set and an executable \spql program, which can be arbitrarily long.   

To increase the number of input cases, we leverage a recently proposed sparse-attention transformer architecture --- \bigb \cite{zaheer2020big}. Instead of having each token attend to all input tokens as in a standard transformer, each token attends to only nearby tokens. Additionally, a small set of global tokens attend to all tokens in the input. This reduces the transformer's memory complexity from quadratic to linear, and empirically, \bigb enables us to use many more cases.

\textbf{Description of input}: The input query $q$ and cases $\mathcal{D}_q = \{(q'_1, \ell'_1), (q'_2, \ell'_2), \ldots (q'_k, \ell'_k)\}$ are concatenated on the encoder side. Specifically, $\mathtt{Input_{ENC}(q, \mathcal{D}_q)} = \mathtt{q} \texttt{[SEP]} \mathtt{q}'_1 \texttt{[SEP]} \mathtt{\ell}'_1, \ldots \mathtt{q_k}' \texttt{[SEP]} \mathtt{\ell_k}'$, where \texttt{[SEP]} denotes the standard separator token. Each logical form also contain the KB entity id of each entities in the question (e.g. m.03\_r3 for Jamaica in Figure~\ref{fig:sparql_example}). We append the entity id after the surface form of the entity mention in the question string. For example, the query in Figure~\ref{fig:sparql_example} becomes "What do Jamaican m.03\_r3 people speak?".

Training is done using a standard seq2seq cross-entropy objective. Large deep neural networks usually benefit from ``good'' initialization points \citep{frankle2019lottery} and being able to utilize pre-trained weights is critical for seq2seq models. We find it helpful to have a regularization term that minimizes the Kullback–Leibler divergence (KLD) between output softmax layers of (1) when only the query $q$ is presented (i.e not using cases), and (2) when query and cases ($\mathcal{D}_q$) are available \cite{yu2013kl}.
Formally, let $f$ be the seq2seq model, let $\sigma = softmax(f(q, \mathcal{D}_q))$ and $\sigma' = softmax(f(q))$ be the decoder's prediction distribution with and without cases, respectively. The following KLD term is added to the seq2seq cross-entropy loss
\begin{equation*}
    L = L_{CE}(f(q, \mathcal{D}_q), l_q) + \lambda_T \textsc{KLD}(\sigma, \sigma')
\end{equation*}
 where $\lambda_T \in [0,1]$ is a hyper-parameter. Intuitively, this term regularizes the prediction of $f(q, \mathcal{D}_q)$ not to deviate too far away from that of the $f(q)$ and we found this to work better than initializing with a model not using cases.

\subsection{Revise}
\label{sub:revise}

In the previous step, the model explicitly reuses the relations present in $\mathcal{D}_q$, nonetheless, there is no guarantee that the query relations in $\mathcal{D}_q$ will contain the relations required to answer the original query $q$. 
This can happen when the domain of $q$ and domain of cases in $\mathcal{D}_q$ are different even when  the relations are semantically similar.  For example, in Figure~\ref{fig:intro2} although the retrieved relations in NN queries are semantically similar, there is a domain mismatch (person v/s fictional characters). Similarly, large KBs are very incomplete \cite{min2013distant}, so querying with a valid relation might require an edge that is missing in the KB leading to intermediate logical forms which do not execute.

To alleviate this problem and to make the queries executable, we explicitly \emph{align} the generated relations with relations (edges) present in the local neighborhood of the query entity in the KG. We propose the following alignment models:

\textbf{Using pre-trained KB embeddings}: KB completion is an extensively studied research field \cite{nickel2011three,bordes2013translating,socher2013reasoning,Velickovic2018GraphAN,sun2019rotate} and several methods have been developed that learn low dimensional representation of relations such that similar relations are closer to each other in the embedding space. We take advantage of the pre-trained relations obtained from TransE \cite{bordes2013translating}, a widely used model for KB completion. For each predicted relation, we find the most similar (outgoing or incoming) relation edge (in terms of cosine similarity) that exists in the KB for that entity and align with it. If the predicted edge exists in the KB, it trivially aligns with itself. There can be multiple missing edges that needs alignment (Figure~\ref{fig:intro2}) and we find it more effective to do beam-search instead of greedy-matching the most similar edge at each step.

\textbf{Using similarity in surface forms}: Similar relations (even across domains) have overlap in their surface forms (e.g. `siblings' is common term in both `person.siblings' and `fictional\_character.siblings'). Therefore, word embeddings obtained by encoding these words should be similar. This observation has been successfully utilized in previous works \cite{toutanova-chen-2015-observed,hwang2019comprehensive}. We similarly encode the predicted relation and all the outgoing or incoming edges with \roberta-base model. Following standard practices, relation strings are prepended with a $\mathtt{[CLS]}$ token and the word pieces are encoded with the \roberta-base model and the output embedding of the $\mathtt{[CLS]}$ token is considered as the relation representation. Similarity between two relation representations is computed by cosine similarity.

Our alignment is simple and requires no learning. By aligning only to individual edges in the KB, we make sure that we do not change the structure of the generated LF. We leave the exploration of learning to align single edges in the program to sequence of edges (paths) in the KB as future work.

\section{Experiments}
\label{sec:experiments}
\begin{table}[t]
    \centering
    \footnotesize
    \setlength{\tabcolsep}{3pt}
    \begin{tabular}{@{}l@{\hskip 0pt}c c c c@{}}
    \toprule
    Model & P & R & F1 & Acc\\\midrule
    \textit{Weakly supervised models} & & & & \\\midrule
    GraftNet \cite{sun2018open} & - & - & 66.4\ensuremath{^\dagger} & - \\
    PullNet \cite{sun2019pullnet} & - & - & 68.1\ensuremath{^\dagger} & - \\
    EmbedKGQA \cite{saxena2020improving} & - & - & 66.6\ensuremath{^\dagger} & - \\
    \midrule
    \textit{Supervised models} & & & & \\\midrule
    STAGG \cite{yih2016value} & 70.9 & \textbf{80.3} & 71.7 & 63.9\\
    T5-11B \cite{t5} &62.1 & 62.6& 61.5& 56.5\\
    T5-11B + Revise &63.6 & 64.3& 63.0& 57.7\\
    \alg (Ours) & \textbf{73.1} & 75.1 & \textbf{72.8} & \textbf{69.9}\\\bottomrule
    
    \end{tabular}
    \caption{Performance on the WebQSP dataset. GraftNet, PullNet and EmbedKGQA produces a ranking of KG entities hence evaluation is in Hits@k (see text for description). \alg significantly outperforms baseline models in the strict exact match accuracy metric. \ensuremath{^\dagger} Models report hits@1 instead of F1}
    \label{tab:webqsp_results}
    \vspace{-3mm}
\end{table}

\noindent\textbf{Data}: For all our experiments, the underlying KB is full Freebase containing over 45 million entities (nodes) and 3 billion facts (edges) \cite{fb}. We test \alg on three datasets --- WebQSP \cite{yih2016value}, CWQ \cite{Talmor2018TheWA} and CFQ \cite{keysers2020measuring}. Please refer to \S\ref{sub:appendix_data} for a detailed description of each datasets.

\noindent\textbf{Hyperparameters}: All hyperparameters are set by tuning on the valdation set for each dataset. We initialize our retriever with the pre-trained \roberta-base weights. We set $p_{\textrm{mask}}=0.2$ for CWQ and 0.5 for the remaining datasets. We use a \bigb generator network with 6 encoding and 6 decoding sparse-attention layers, which we initialize with pre-trained \textsc{Bart}-base weights. We use $k$=20 cases and decode with a beam size of 5. Initial learning rate is set to $5\times 10^{-5}$ and is decayed linearly through training. Further details for the EMNLP reproducibility checklist is given in \S\ref{sub:appendix_hyperparam}.

\subsection{Entity Linking}
\label{sub:entity_linking}
The first step required to generate an executable LF for a NL query is to identify and link the entities present in the query. For our experiments, we use a combination of an off-the-shelf entity linker and a large mapping of mentions to surface forms. For the off-the-shelf linker, we use a recently proposed high precision entity linker \textsc{Elq} \cite{li2020efficient}. To further improve recall of our system, we first identify mention spans of entities in the question by tagging it with a NER\footnote{\url{https://cloud.google.com/natural-language}} system. Next, we link entities not linked by \textsc{Elq} by exact matching with surface form annotated in FACC1 project \cite{facc1}. Our entity linking results are shown in Table~\ref{tab:el_results}.
\begin{table}[]
    \centering
    \small
    \begin{tabular}{@{}l c c c@{}}
    \toprule
    Dataset &  Precision & Recall & F1\\\midrule
    WebQSP  & 0.761   & 0.819 & 0.789\\
    CWQ & 0.707 & 0.910 & 0.796\\\bottomrule
    \end{tabular}
    \caption{Entity linking performance on various datasets}
    \label{tab:el_results}
    \vspace{-1em}
\end{table}

\subsection{KBQA Results}
\label{sub:main_results}
This section reports the performance of \alg on various benchmarks. We report the strict exact match accuracy where we compare the list of predicted answers by executing the generated \spql program to the list of gold answers\footnote{We generate the gold answer entities by executing the gold \spql query against our Freebase KB}. A question is answered correctly if the two list match exactly. We also report the precision, recall and the F1 score to be comparable to the baselines. Models such as GraftNet \cite{sun2018open} and PullNet \cite{sun2019pullnet} rank answer entities and return the top entity as answer (Hits@1 in table~\ref{tab:webqsp_results}). This is undesirable for questions that have multiple entities as answers (e.g. ``Name the countries bordering the U.S.?''). We also report performance of models that only depend on the query and answer pair during training and do not depend on LF supervision (weakly-supervised setting). Unsurprisingly, models trained with explicit LF supervision perform better than weakly supervised models. Our main baseline is a massive pre-trained seq2seq model with orders of magnitude more number of parameters --- T5-11B \cite{t5}. T5 has recently been shown to be effective for compositional KBQA \cite{furrer2020compositional}. For each dataset, we fine-tune the T5 model on the query and the LF pairs. 

\begin{table}[t]
\centering
\footnotesize
\setlength{\tabcolsep}{4pt}
\begin{tabular}{@{}l c c c c@{}}
\toprule
Model & P & R & F1 & Acc \\
\midrule
\textit{Weakly supervised models} & & & & \\\midrule
KBQA-GST \cite{lan2019knowledge} & - & - & - & 39.4\\
QGG  \cite{lan2020query} & - & - & - & 44.1 \\
PullNet \cite{sun2019pullnet}& - & - & - & 45.9 \\
DynAS (Alibaba Group) & - & - & - & 50.0 \\
\midrule
\textit{Supervised models} & & & & \\\midrule
T5-11B \cite{t5} & 55.2 & 55.4 & 54.6 & 52.4\\
T5-11B + Revise & 58.7 & 59.6& 58.2& 55.6\\
\alg (Ours) & \textbf{70.4} & \textbf{71.9} & \textbf{70.0} & \textbf{67.1} \\
\bottomrule
\end{tabular}
\caption{Performance on the hidden test set of CWQ.} 
\label{tab:cwq_results}
\end{table}

Table~\ref{tab:webqsp_results} reports results of various models on WebQSP. 
All reported model except \alg and T5-11B directly operate on the KB (e.g. traverse KB paths starting from the query entity) to generate the LF or the answer. As a result, models such as STAGG tend to enjoy much higher recall. On the other hand, much of our logical query is generated by reusing components of similar cases. We also report the results of `aligning' the LF produced by T5 using our revise step.  As shown in Table~\ref{tab:webqsp_results}, \alg outperforms all other models significantly and improves on the strict exact-match accuracy by more than 6 points w.r.t. the best model. Revise step also improves on the performance of T5 suggesting that it is generally applicable. Table~\ref{tab:cwq_results} reports performance on the hidden test set of CWQ\footnote{The result of our model in the official leaderboard (\url{https://www.tau-nlp.org/compwebq-leaderboard}) is higher (70.4 vs 67.1). This is because the official evaluation script assigns full score if any of the correct answer entities are returned even if there are multiple correct answers for a question. In the paper we report strict exact-match accuracy.}, which was built by extending WebQSP questions with the goal of making a more complex multi-hop questions. It is encouraging to see that \alg outperforms all other baselines on this challenging dataset by a significant margin. Finally, we report results on CFQ in Table~\ref{tab:my_label}. On error analysis, we found that on several questions which are yes/no type, our model was predicting the list of correct entities instead of predicting a yes or no. We created a rule-based binary classifier that predicts the type of question (yes/no or other). If the question was predicted as a yes/no, we would output a yes if the length of the predicted answer list was greater than zero and no otherwise. (If the model was already predicting a yes/no, we keep the original answer unchanged). We report results on all the three MCD splits of the dataset and compare with the T5-11B model of \citet{furrer2020compositional} and we find that our model outperforms T5-11B on this dataset as well. It is encouraging to see that \alg, even though containing order-of-magnitudes less parameters than T5-11B, outperforms it on all benchmarks showing that it is possible for smaller models with less carbon footprint and added reasoning capabilities to outperform massive pre-trained LMs.

\begin{table}[t]
    \centering
    \small
    \begin{tabular}{@{}l c c c c@{}}
    \toprule
        Model  & MCD1 & MCD2 & MCD3 & MCD-mean\\\midrule
         T5-11B & 72.9 & 69.2 & 62.0 & 67.7\\
         \alg & \textbf{87.8} & \textbf{75.1} & \textbf{71.5} & \textbf{78.1}\\
    \bottomrule
    \end{tabular}
    \caption{Performance (accuracy) on the CFQ dataset.}
    \label{tab:my_label}
    \vspace{-5mm}
\end{table}


\subsection{Efficacy of Revise step}
\label{sub:revise_step}
Table~\ref{tab:revise} show that the revise step is useful for \alg on multiple datasets. We also show that the T5 model also benefits from the alignment in revise step with more than 3 points improvement in F1 score on the CWQ dataset. We find that TransE alignment outperforms \roberta based alignment, suggesting that graph structure information is more useful than surface form similarity for aligning relations. Moreover, relation names are usually short strings, so they do not provide enough context for LMs to form good representations.

Next we demonstrate the advantage of the nonparametric property of \alg --- ability to fix an initial wrong prediction by allowing new cases to be \emph{injected} to the case-memory. This allows \alg to generalize to queries which needs relation \emph{never seen} during training. Due to space constraints, we report other results (e.g. retriever performance), ablations and other analysis in \S\ref{sec:appendix_further_experiments}.

\subsection{Point-Fixes to Model Predictions}
\label{sub:analysis}
Modern QA systems built on top of large LMs do not provide us the opportunity to debug an erroneous prediction. The current approach is to fine-tune the model on new data. However, this process is time-consuming and impractical for production settings. Moreover, it has been shown (and as we will empirically demonstrate) that this approach leads to catastrophic forgetting where the model forgets what it had learned before. \cite{mccloskey1989catastrophic,kirkpatrick2017overcoming}. On the other hand, \alg adopts a nonparametric approach and allows inspection of the retrieved nearest neighbors for a query. Moreover, one could \emph{inject} a new relevant case into the case-memory (KNN-index), which could be picked up by the retriever and used by the reuse module to fix an erroneous prediction.

\begin{table}[t]
    \centering
    \small
    \begin{tabular}{@{}l c c@{}}
    \toprule
        WebQSP &  Accuracy(\%) & $\Delta$\\
        \midrule
        \alg (before Revise) & 69.43 & --\\
        $\quad+$Revise (Roberta) & 69.49 & +0.06\\
        $\quad+$Revise (TransE) & \textbf{70.00} & \textbf{+0.57}\\
        \midrule
        CWQ &  Accuracy(\%) & $\Delta$\\
        \midrule
        \alg (before Revise) & 65.95 & --\\
        $\quad+$Revise (Roberta) & 66.32 & +0.37 \\
        $\quad+$Revise (TransE) & \textbf{67.11} & \textbf{+1.16}\\
    \bottomrule
    \end{tabular}
    \caption{Impacts of the revise step. We show that the revise step consistently improves the accuracy on WebQSP and CWQ, especially with the TransE pretrained embeddings.}
    \label{tab:revise}
    \vspace{-6mm}
\end{table}

\begin{table*}
    \centering
    \small
    \begin{tabular}{@{}l c c c@{}}
    \toprule
    Scenario && Initial Set & Held-Out \\
    \midrule
    Transformer     && 59.6 & 0.0 \\
    \quad + Fine-tune on additional cases only (100 gradient steps) && 1.3 & 76.3 \\
    \quad + Fine-tune on additional cases and original data (300 gradient steps) && 53.1 & 57.6 \\
    \midrule
    \alg (Ours) && 69.4 & 0.0 \\
    \quad + Adding additional cases to index (0 gradient steps; 2 sec) && 69.4 & 70.6 \\
    \bottomrule
    \end{tabular}
    \caption{\small Robustness and controllability of our method against black-box transformers. \alg can easily and quickly adopt to new relations given cases about it, whereas heavily parameterized transformer is not stable, and can undergo catastrophic forgetting when we try to add new relation information intro its parameters.}
    \label{tab:robust}
\end{table*}
\vspace{-2mm}
\begin{table}
    \centering
    \small
    \footnotesize
    \begin{tabular}{@{}l c c c c c@{}}
    \toprule
    Scenario && P & R & F1 & Acc \\
    \midrule
    \alg (Ours) && 0.0 & 0.0 & 0.0 & 0.0 \\
    \quad + additional cases &&  36.54 & 38.59 & 36.39 & 32.89 \\
    \bottomrule
    \end{tabular}
    \caption{\small Results for \hitl experiment. After adding a few cases, we see that we can get the accuracy of OOV questions to improve considerably, without needing to re-train the model.}
    \label{tab:hitl_results_appendix}
    \vspace{-2mm}
\end{table}

\subsubsection{Performance on Unseen Relations}
\label{subsub:unseen_rel}
We consider the case when the model generates a wrong LF for a given query. We create a controlled setup by removing all queries from the training set of WebQSP which contain the (people.person.education) relation. This led to a removal of 136 queries from the train set and ensured that the model failed to correctly answer the 86 queries (held-out) in the test set which contained the removed relation in its LF.

We compare to a baseline transformer model (which do not use cases) as our baseline. As shown in Table~\ref{tab:robust}, both baseline and \alg do not perform well without any relevant cases since a required KB relation was missing during training. Next, we add the 136 training instances back to the training set and recompute the KNN index. This process involves encoding the newly added NL queries and recomputing the KNN index, a process which is computationally much cheaper than re-training the model again. Row 5 in Table~\ref{tab:robust} shows the new result. On addition of the new cases, \alg can seamlessly use them and copy the unseen relation to predict the correct LF, reaching 70.6\% accuracy on the 86 held-out queries. 

\begin{figure}
    \centering
    \includegraphics[width=\columnwidth]{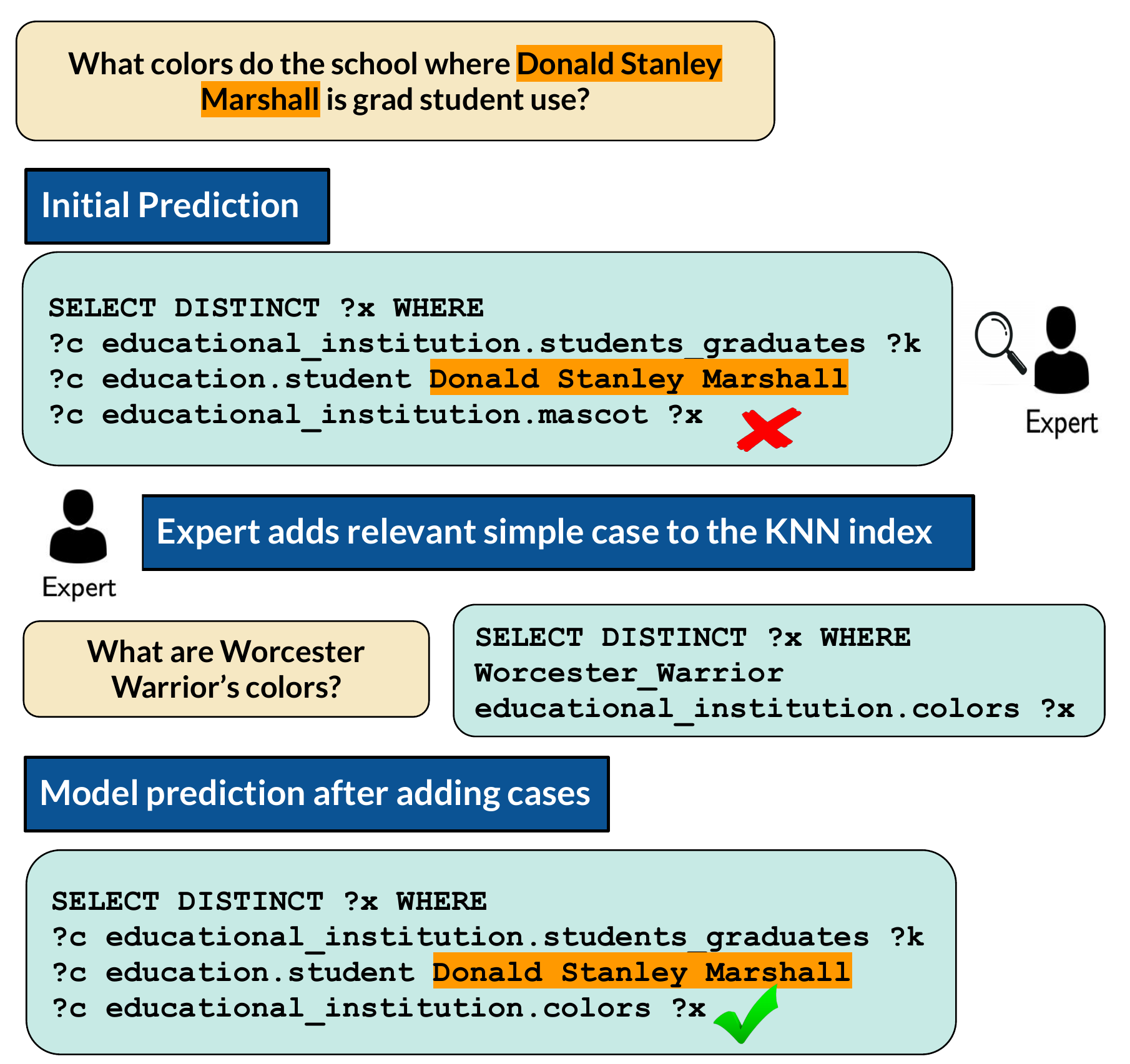}
    \vspace{-5mm}
    \caption{\small An expert point-fixes a model prediction by adding a simple case to the KNN index. Initial prediction was incorrect as no query with the relation (educational\_institution.colors) was present in the train set. \alg retrieves the case from the KNN index and fixes the erroneous prediction without requiring any re-training.}
    \label{fig:hitl_fig}
\end{figure}

In contrast, the baseline transformer must be fine-tuned on the new cases to handle the new relation, which is computationally more expensive than adding the cases to our index. Moreover, just fine-tuning on the new instances leads to \emph{catastrophic forgetting} as seen in row 2 of Table~\ref{tab:robust} where the baseline model's performance on the initial set decreases drastically. We find it necessary to carefully fine-tune the model on new examples alongside original training examples (in a 1:2 proportion). However, it still converges to a performance which is lower than its original performance and much lower than the performance of \alg.

\subsubsection{Human-in-the-Loop Experiment}
\label{subsub:hitl}
During error analysis, we realized that there are queries in the test set of WebQSP that contain KB relations in their LFs which were never seen during training\footnote{There are 94 different unseen relations in test set.}. That means models will never be able to predict the correct LF for the query because of the unseen relation. We conduct a human-in-the-loop experiment (Figure~\ref{fig:hitl_fig}) in which users add \emph{simple} `cases' to point-fix erroneous predictions of \alg for those queries. A simple case is a NL query paired with a program which only contain one KB relation. Table~\ref{tab:hitl_appendix} (Appendix \ref{app:hitl}) shows some example of such cases. Because of the simple nature of the questions, these cases can be created manually (by a user who is knowledgeable about the KB schema) or automatically curated from data sources such as SimpleQuestions \cite{bordes2015large} which is a large collection of NL queries that can be a mapped to a single KB edge. Table~\ref{tab:hilt_stats_appendix} in Appendix \ref{app:hitl} shows various statistics of the missing relations and the number of cases added by humans and from SimpleQuestions. The cases are added to the original KNN-index. By adding a few cases, the performance increases from 0 to 36 F1 (Table~\ref{tab:hitl_results_appendix}) without requiring any training. Note unlike the previous controlled experiment in \S\ref{subsub:unseen_rel}, we add around 3.87 cases for each unseen relation\footnote{In \S\ref{subsub:unseen_rel}, we added 136 cases (v/s 3.87) for one relation. This is why the accuracy in Table~\ref{tab:robust} is higher w.r.t Table~\ref{tab:hitl_results_appendix}.}.\\
\noindent\textbf{Importance of this result}: We believe that flexibility of models to \emph{fix} predictions (without training) is an important desideratum for QA systems deployed in production settings and we hope our results will inspire future research in this direction.
\vspace{-1mm}

\subsection{Further Analysis}
\label{sub:analysis}
\begin{table}[]
\vspace{-1mm}
    \centering
    \small
    \begin{tabular}{@{}l  r c  r r@{}}\toprule
        \multirow{2}{*}{Data} & \multirow{2}{*}{\# Total Q} &
        \multirow{2}{*}{\begin{tabular}[c]{@{}l@{}}\#\, Q that need\\comp. reasoning\end{tabular}}
        & \multicolumn{2}{c}{\# Correct}\\
        \cmidrule{4-5}
        & &  & T5 & CBR \\
        \midrule
        CWQ & 3531 & 639 & 205 & \textbf{270}\\
        CFQ & 11968 & 6541 & 3351 & \textbf{3886}\\\bottomrule
    \end{tabular}
    \vspace{-2mm}
    \caption{\small We compare the performance of models on questions that need \emph{novel combinations} of relations \emph{not seen} during training.}
    \vspace{-6mm}
    \label{tab:comp}
\end{table}

We analyze questions in the evaluation set which require \emph{novel combinations} of relations \emph{never seen} in the training set. This means, in order for our model to answer these questions correctly, it would have to retrieve relevant nearest neighbor (NN) questions from the training set and copy the required relations from  the logical form of \emph{multiple} NN queries. Table~\ref{tab:comp} shows that our model outperforms the competitive T5 baseline. Also as we saw in the last section, our model is able to quickly adapt to relations \emph{never seen} in the training set altogether by picking them up from newly added cases.
\begin{table}[]
\vspace{-4mm}
    \small
    \centering
    \begin{tabular}{l c c}\toprule
          & WebQSP & CWQ  \\\midrule
         Baseline (K = 0) & 67.2 & 65.8 \\
         \alg (K = 20) & \textbf{69.9} & \textbf{67.1} \\
         \quad - KL term in loss & 68.1 & 66.7\\\bottomrule
    \end{tabular}
    \caption{Ablation experiment with a baseline model that do not use cases and also when the KL divergence term (\S2.2) is not used in loss function of reuse step . The numbers denote exact match accuracy.}
    \label{tab:ablation_results_k=0}
\end{table}

\begin{table}[]
\vspace{-3.5mm}
    \centering
    \small
    \begin{tabular}{l c}\toprule
       \# nearest neighbors  &  Accuracy \\\midrule
    K = 0 & 67.20 \\
    K = 1 & 68.45 \\
    K = 10 & 69.23\\
    K = 20 & 69.98\\\bottomrule
    \end{tabular}
    \caption{Performance on WebQSP on varying number of nearest neigbors}
    \vspace{-6mm}
    \label{tab:appendix_vary_k}
\end{table}

We also compare with a model with the same reuse component of \alg but is trained and tested without retrieving any cases from the case-memory (Table~\ref{tab:ablation_results_k=0}). Even though the baseline model is competitive, having similar cases is beneficial, especially for the WebQSP dataset. We also report the results when we only use cross-entropy loss for training the \bigb model and not the KL-divergence term. Table~\ref{tab:appendix_vary_k} reports the performance of \alg using different number of retrieved cases. It is encouraging to see the the performance improves with increasing number of cases.

\section{Related Work}
\vspace{-2mm}
\label{sec:related_work}

\textbf{Retrieval augmented QA models} ~\cite{chen2017reading,guu2020realm,lewis2020retrieval} augments a reader model with a retriever to find relevant paragraphs from a nonparametric memory. In contrast, our CBR approach retrieves similar queries and uses their logical forms to derive a new solution. Recently \citet{lewis2020question} proposed a model that finds a nearest neighbor (NN) question in the training set and returns the answer to that question. While this model would be helpful if the exact question or its paraphrase is present in the training set, it will not generalize to other scenarios. \alg, on the other hand, learns to reason with the retrieved programs of multiple retrieved NN queries and generates a new program for the given query and hence is able to generalize even if the query paraphrase is not present in the train set.\\
\noindent\textbf{Retrieve and edit}: \alg shares similarities with the \textsc{retrieve-and-edit} framework \cite{hashimoto2018retrieve} which utilizes retrieved nearest neighbor for structured prediction. However, unlike our method they only retrieve a single nearest neighbor and will unlikely be able to generate programs for questions requiring relations from multiple nearest neighbors.\\
\noindent{\textbf{Generalizing to unseen database schemas}:} There has been work in program synthesis that generates SQL programs for unseen database schemas \cite{rat-sql,lin2020bridging}. However, these work operate on web or Wikipedia tables with small schemas. For example, in WikiTableQuestions \cite{pasupat2015compositional} the average. number of columns in a table is 5.8 and in Spider dataset \cite{yu2018spider}, the average number of columns is 28.1. On the other hand, our model has to consider all possible Freebase relations (in thousands). Previous work perform schema-aware encoding which is not possible in our case because of the large number of relations. The retrieve step of \alg can be seen as a pruning step which narrows the number of candidate relations by retrieving relevant questions and their logical forms.\\
\noindent\textbf{Case-based Reasoning for KB completion}: Recently, a CBR based KB reasoning approach was proposed by \citet{das2020simple,das-etal-2020-probabilistic}. They retrieve similar entities and then find KB reasoning paths from them. However, their approach does not handle complex natural language queries and only operate on structured triple queries. Additionally, the logical forms handled by our model have much more expressive power than knowledge base paths.\\
\noindent\textbf{Program Synthesis and Repair}: Repairing / revising generated programs has been studied in the field of program synthesis. For example, prior work repairs a program based on syntax of the underlying language \cite{le2017s3}, by generating sketches \cite{hua2018towards}. More recently, \citet{gupta2020synthesize} proposes a framework in which they use a program debugger to revise the program generated by a neural program synthesizer. However, none of these works take advantage of the similarity between semantic relations present in the knowledge base, and hence, unlike us, they do not use embeddings of similar relation to align relations. More generally, many prior efforts have employed neural models to generate \spql -like code for semantic parsing \cite{dong2016language,balog2016deepcoder,zhong2017seq2sql}, SQL queries over relational databases~\cite{zhongSeq2SQL2017}, program-structured neural network layouts~\cite{Andreas_2016_CVPR}, or even proofs for mathematical theorems~\cite{polu2020generative}.
Our work differs in our use of the programs of multiple retrieved similar queries to generate the target program.

\noindent\textbf{K-NN approach in other NLP applications}:  \citet{khandelwal2019generalization} demonstrate improvements in language modeling by utilizing explicit examples from training data. There has been work in machine translation \cite{zhang2018guiding,gu2018search,khandelwal2020nearest} that uses nearest neighbor translation pair to guide the decoding process. Recently, \citet{hossain2020simple} proposed a retrieve-edit-rerank approach for text generation in which each retrieved candidate from the training set is edited independently and then re-ranked. In contrast, \alg generates the program \emph{jointly} from all the retrieved cases and is more suitable for questions which needs copying relations from multiple nearest neighbors. Please refer to (\S\ref{sec:further_related_work}) for further related work.

\section{Limitations and Future Work}
\vspace{-2mm}
\label{sec:limitations}
To the best of our knowledge, we are the first to propose a neuralized CBR approach for KBQA. We showed that our model is effective in handling complex questions over KBs, but our work also has several limitations. First, our model relies on the availability of supervised logical forms such as \spql queries, which can be expensive to annotate at scale. In the future, we plan to explore ways to directly learn from question-answer pairs \cite{berant2013semantic,liang2016neural}. Even though, \alg is modular and has several advantages, the retrieve and reuse components of our model are trained separately. In future, we plan to explore avenues for end to end learning for CBR.

\section*{Acknowledgments}
We thank Prof. Yu Su  and Yu Gu (Ohio State University) for their help in setting up the Freebase server, Andrew Drozdov, Kalpesh Krishna, Subendhu Rongali and other members of the
UMass IESL and NLP groups for helpful discussion and feedback. RD and DT are funded in part by the Chan Zuckerberg Initiative under the project Scientific Knowledge Base Construction and in part by IBM Congitive Horizons Network (CHN).
EP is grateful for support from the NSF Graduate Research Fellowship and the Open Philanthropy AI Fellowship. The work reported here was performed in part by the Center for Data Science and the Center for Intelligent Information Retrieval, and in part using high performance computing equipment obtained under a grant from the Collaborative R\&D Fund managed by the Massachusetts Technology Collaborative.

\bibliography{anthology,custom}

\begin{thebibliography}{87}
\expandafter\ifx\csname natexlab\endcsname\relax\def\natexlab#1{#1}\fi

\bibitem[{Aamodt and Plaza(1994)}]{aamodt1994case}
Agnar Aamodt and Enric Plaza. 1994.
\newblock Case-based reasoning: Foundational issues, methodological variations,
  and system approaches.
\newblock \emph{AI communications}.

\bibitem[{Andreas et~al.(2016)Andreas, Rohrbach, Darrell, and
  Klein}]{Andreas_2016_CVPR}
Jacob Andreas, Marcus Rohrbach, Trevor Darrell, and Dan Klein. 2016.
\newblock Neural module networks.
\newblock In \emph{Proceedings of the IEEE Conference on Computer Vision and
  Pattern Recognition (CVPR)}.

\bibitem[{Balog et~al.(2016)Balog, Gaunt, Brockschmidt, Nowozin, and
  Tarlow}]{balog2016deepcoder}
Matej Balog, Alexander~L Gaunt, Marc Brockschmidt, Sebastian Nowozin, and
  Daniel Tarlow. 2016.
\newblock Deepcoder: Learning to write programs.
\newblock \emph{arXiv preprint arXiv:1611.01989}.

\bibitem[{Berant et~al.(2013)Berant, Chou, Frostig, and
  Liang}]{berant2013semantic}
Jonathan Berant, Andrew Chou, Roy Frostig, and Percy Liang. 2013.
\newblock Semantic parsing on freebase from question-answer pairs.
\newblock In \emph{EMNLP}.

\bibitem[{Bollacker et~al.(2008)Bollacker, Evans, Paritosh, Sturge, and
  Taylor}]{fb}
Kurt Bollacker, Colin Evans, Praveen Paritosh, Tim Sturge, and Jamie Taylor.
  2008.
\newblock Freebase: A collaboratively created graph database for structuring
  human knowledge.
\newblock In \emph{ICDM}.

\bibitem[{Bordes et~al.(2015)Bordes, Usunier, Chopra, and
  Weston}]{bordes2015large}
Antoine Bordes, Nicolas Usunier, Sumit Chopra, and Jason Weston. 2015.
\newblock Large-scale simple question answering with memory networks.
\newblock \emph{arXiv preprint arXiv:1506.02075}.

\bibitem[{Bordes et~al.(2013)Bordes, Usunier, Garcia-Duran, Weston, and
  Yakhnenko}]{bordes2013translating}
Antoine Bordes, Nicolas Usunier, Alberto Garcia-Duran, Jason Weston, and Oksana
  Yakhnenko. 2013.
\newblock Translating embeddings for modeling multi-relational data.
\newblock In \emph{Neurips}.

\bibitem[{Cao et~al.(2018)Cao, Li, Li, and Wei}]{cao-etal-2018-retrieve}
Ziqiang Cao, Wenjie Li, Sujian Li, and Furu Wei. 2018.
\newblock Retrieve, rerank and rewrite: Soft template based neural
  summarization.
\newblock In \emph{ACL}.

\bibitem[{Chen et~al.(2017)Chen, Fisch, Weston, and Bordes}]{chen2017reading}
Danqi Chen, Adam Fisch, Jason Weston, and Antoine Bordes. 2017.
\newblock Reading wikipedia to answer open-domain questions.
\newblock In \emph{ACL}.

\bibitem[{Daelemans et~al.(1996)Daelemans, Zavrel, Berck, and
  Gillis}]{daelemans1996mbt}
Walter Daelemans, Jakub Zavrel, Peter Berck, and Steven Gillis. 1996.
\newblock Mbt: A memory-based part of speech tagger-generator.
\newblock In \emph{WVLC}.

\bibitem[{Das et~al.(2019)Das, Dhuliawala, Zaheer, and McCallum}]{das2019multi}
Rajarshi Das, Shehzaad Dhuliawala, Manzil Zaheer, and Andrew McCallum. 2019.
\newblock Multi-step retriever-reader interaction for scalable open-domain
  question answering.
\newblock In \emph{ICLR}.

\bibitem[{Das et~al.(2020{\natexlab{a}})Das, Godbole, Dhuliawala, Zaheer, and
  McCallum}]{das2020simple}
Rajarshi Das, Ameya Godbole, Shehzaad Dhuliawala, Manzil Zaheer, and Andrew
  McCallum. 2020{\natexlab{a}}.
\newblock A simple approach to case-based reasoning in knowledge bases.
\newblock In \emph{AKBC}.

\bibitem[{Das et~al.(2020{\natexlab{b}})Das, Godbole, Monath, Zaheer, and
  McCallum}]{das-etal-2020-probabilistic}
Rajarshi Das, Ameya Godbole, Nicholas Monath, Manzil Zaheer, and Andrew
  McCallum. 2020{\natexlab{b}}.
\newblock Probabilistic case-based reasoning for open-world knowledge graph
  completion.
\newblock In \emph{Findings of EMNLP}.

\bibitem[{Dong and Lapata(2016)}]{dong2016language}
Li~Dong and Mirella Lapata. 2016.
\newblock Language to logical form with neural attention.
\newblock In \emph{ACL}.

\bibitem[{Ferrucci et~al.(2010)Ferrucci, Brown, Chu-Carroll, Fan, Gondek,
  Kalyanpur, Lally, Murdock, Nyberg, Prager, Schlaefer, and
  Welty}]{ferrucci2010building}
David Ferrucci, Eric Brown, Jennifer Chu-Carroll, James Fan, David Gondek,
  Aditya~A. Kalyanpur, Adam Lally, J.~William Murdock, Eric Nyberg, John
  Prager, Nico Schlaefer, and Chris Welty. 2010.
\newblock \href {https://doi.org/10.1609/aimag.v31i3.2303} {Building watson: An
  overview of the deepqa project}.
\newblock \emph{AI Magazine}, 31(3):59--79.

\bibitem[{Frankle and Carbin(2019)}]{frankle2019lottery}
Jonathan Frankle and Michael Carbin. 2019.
\newblock \href {http://arxiv.org/abs/1803.03635} {The lottery ticket
  hypothesis: Finding sparse, trainable neural networks}.
\newblock In \emph{ICLR}.

\bibitem[{Furrer et~al.(2020)Furrer, van Zee, Scales, and
  Sch{\"a}rli}]{furrer2020compositional}
Daniel Furrer, Marc van Zee, Nathan Scales, and Nathanael Sch{\"a}rli. 2020.
\newblock Compositional generalization in semantic parsing: Pre-training vs.
  specialized architectures.
\newblock \emph{arXiv preprint arXiv:2007.08970}.

\bibitem[{Gabrilovich et~al.(2013)Gabrilovich, Ringgaard, and
  Subramanya}]{facc1}
Evgeniy Gabrilovich, Michael Ringgaard, and Amarnag Subramanya. 2013.
\newblock Facc1: Freebase annotation of clueweb corpora, version 1 (release
  date 2013-06-26, format version 1, correction level 0).

\bibitem[{Gu et~al.(2018)Gu, Wang, Cho, and Li}]{gu2018search}
Jiatao Gu, Yong Wang, Kyunghyun Cho, and Victor~OK Li. 2018.
\newblock Search engine guided neural machine translation.
\newblock In \emph{AAAI}.

\bibitem[{Gupta et~al.(2020)Gupta, Christensen, Chen, and
  Song}]{gupta2020synthesize}
Kavi Gupta, Peter~Ebert Christensen, Xinyun Chen, and Dawn Song. 2020.
\newblock Synthesize, execute and debug: Learning to repair for neural program
  synthesis.
\newblock In \emph{Neurips}.

\bibitem[{Guu et~al.(2020)Guu, Lee, Tung, Pasupat, and Chang}]{guu2020realm}
Kelvin Guu, Kenton Lee, Zora Tung, Panupong Pasupat, and Ming-Wei Chang. 2020.
\newblock Realm: Retrieval-augmented language model pre-training.
\newblock In \emph{ICML}.

\bibitem[{Hashimoto et~al.(2018)Hashimoto, Guu, Oren, and
  Liang}]{hashimoto2018retrieve}
Tatsunori~B Hashimoto, Kelvin Guu, Yonatan Oren, and Percy Liang. 2018.
\newblock A retrieve-and-edit framework for predicting structured outputs.
\newblock In \emph{Neurips}.

\bibitem[{Hinton and Plaut(1987)}]{hinton1987using}
Geoffrey~E Hinton and David~C Plaut. 1987.
\newblock Using fast weights to deblur old memories.
\newblock In \emph{Proceedings of the ninth annual conference of the Cognitive
  Science Society}.

\bibitem[{Hossain et~al.(2020)Hossain, Ghazvininejad, and
  Zettlemoyer}]{hossain2020simple}
Nabil Hossain, Marjan Ghazvininejad, and Luke Zettlemoyer. 2020.
\newblock Simple and effective retrieve-edit-rerank text generation.
\newblock In \emph{ACL}.

\bibitem[{Hua et~al.(2018)Hua, Zhang, Wang, and Khurshid}]{hua2018towards}
Jinru Hua, Mengshi Zhang, Kaiyuan Wang, and Sarfraz Khurshid. 2018.
\newblock Towards practical program repair with on-demand candidate generation.
\newblock In \emph{international conference on software engineering}, pages
  12--23.

\bibitem[{Hua et~al.(2020)Hua, Li, Haffari, Qi, and Wu}]{hua2020retrieve}
Yuncheng Hua, Yuan-Fang Li, Reza Haffari, Guilin Qi, and Wei Wu. 2020.
\newblock Retrieve, program, repeat: complex knowledge base question answering
  via alternate meta-learning.
\newblock In \emph{International Joint Conference on Artificial Intelligence
  2020}, pages 3679--3686. Association for the Advancement of Artificial
  Intelligence (AAAI).

\bibitem[{Hwang et~al.(2019)Hwang, Yim, Park, and Seo}]{hwang2019comprehensive}
Wonseok Hwang, Jinyeong Yim, Seunghyun Park, and Minjoon Seo. 2019.
\newblock A comprehensive exploration on wikisql with table-aware word
  contextualization.
\newblock \emph{arXiv preprint arXiv:1902.01069}.

\bibitem[{Karpukhin et~al.(2020)Karpukhin, O{\u{g}}uz, Min, Wu, Edunov, Chen,
  and Yih}]{karpukhin2020dense}
Vladimir Karpukhin, Barlas O{\u{g}}uz, Sewon Min, Ledell Wu, Sergey Edunov,
  Danqi Chen, and Wen-tau Yih. 2020.
\newblock Dense passage retrieval for open-domain question answering.
\newblock In \emph{EMNLP}.

\bibitem[{Keysers et~al.(2020)Keysers, Sch{\"a}rli, Scales, Buisman, Furrer,
  Kashubin, Momchev, Sinopalnikov, Stafiniak, Tihon, Tsarkov, Wang, van Zee,
  and Bousquet}]{keysers2020measuring}
Daniel Keysers, Nathanael Sch{\"a}rli, Nathan Scales, Hylke Buisman, Daniel
  Furrer, Sergii Kashubin, Nikola Momchev, Danila Sinopalnikov, Lukasz
  Stafiniak, Tibor Tihon, Dmitry Tsarkov, Xiao Wang, Marc van Zee, and Olivier
  Bousquet. 2020.
\newblock Measuring compositional generalization: A comprehensive method on
  realistic data.
\newblock In \emph{ICLR}.

\bibitem[{Khandelwal et~al.(2021)Khandelwal, Fan, Jurafsky, Zettlemoyer, and
  Lewis}]{khandelwal2020nearest}
Urvashi Khandelwal, Angela Fan, Dan Jurafsky, Luke Zettlemoyer, and Mike Lewis.
  2021.
\newblock Nearest neighbor machine translation.
\newblock In \emph{ICLR}.

\bibitem[{Khandelwal et~al.(2020)Khandelwal, Levy, Jurafsky, Zettlemoyer, and
  Lewis}]{khandelwal2019generalization}
Urvashi Khandelwal, Omer Levy, Dan Jurafsky, Luke Zettlemoyer, and Mike Lewis.
  2020.
\newblock Generalization through memorization: Nearest neighbor language
  models.
\newblock In \emph{ICLR}.

\bibitem[{Kirkpatrick et~al.(2017)Kirkpatrick, Pascanu, Rabinowitz, Veness,
  Desjardins, Rusu, Milan, Quan, Ramalho, Grabska-Barwinska
  et~al.}]{kirkpatrick2017overcoming}
James Kirkpatrick, Razvan Pascanu, Neil Rabinowitz, Joel Veness, Guillaume
  Desjardins, Andrei~A Rusu, Kieran Milan, John Quan, Tiago Ramalho, Agnieszka
  Grabska-Barwinska, et~al. 2017.
\newblock Overcoming catastrophic forgetting in neural networks.
\newblock \emph{PNAS}.

\bibitem[{Kolodner(1983)}]{kolodner1983maintaining}
Janet~L Kolodner. 1983.
\newblock Maintaining organization in a dynamic long-term memory.
\newblock \emph{Cognitive science}.

\bibitem[{Lake and Baroni(2018)}]{lake_baroni}
Brenden~M Lake and Marco Baroni. 2018.
\newblock Generalization without systematicity.
\newblock In \emph{ICML}.

\bibitem[{Lan and Jiang(2020)}]{lan2020query}
Yunshi Lan and Jing Jiang. 2020.
\newblock Query graph generation for answering multi-hop complex questions from
  knowledge bases.
\newblock In \emph{ACL}.

\bibitem[{Lan et~al.(2019)Lan, Wang, and Jiang}]{lan2019knowledge}
Yunshi Lan, Shuohang Wang, and Jing Jiang. 2019.
\newblock Knowledge base question answering with topic units.
\newblock In \emph{IJCAI}.

\bibitem[{Lancaster and Kolodner(1987)}]{lancaster1987problem}
Juliana~S Lancaster and Janet~L Kolodner. 1987.
\newblock Problem solving in a natural task as a function of experience.
\newblock Technical report, Georgia Tech CS Department.

\bibitem[{Le et~al.(2017)Le, Chu, Lo, Le~Goues, and Visser}]{le2017s3}
Xuan-Bach~D Le, Duc-Hiep Chu, David Lo, Claire Le~Goues, and Willem Visser.
  2017.
\newblock S3: syntax-and semantic-guided repair synthesis via programming by
  examples.
\newblock In \emph{Foundations of Software Engineering}.

\bibitem[{Leake(1996)}]{leake1996cbr}
David~B Leake. 1996.
\newblock Cbr in context: The present and future.
\newblock \emph{Case-based reasoning: Experiences, lessons, and future
  directions}.

\bibitem[{LeCun et~al.(2015)LeCun, Bengio, and Hinton}]{lecun2015deep}
Yann LeCun, Yoshua Bengio, and Geoffrey Hinton. 2015.
\newblock Deep learning.
\newblock \emph{nature}.

\bibitem[{Lewis et~al.(2020{\natexlab{a}})Lewis, Liu, Goyal, Ghazvininejad,
  Mohamed, Levy, Stoyanov, and Zettlemoyer}]{lewis-etal-2020-bart}
Mike Lewis, Yinhan Liu, Naman Goyal, Marjan Ghazvininejad, Abdelrahman Mohamed,
  Omer Levy, Veselin Stoyanov, and Luke Zettlemoyer. 2020{\natexlab{a}}.
\newblock {BART}: Denoising sequence-to-sequence pre-training for natural
  language generation, translation, and comprehension.
\newblock In \emph{ACL}.

\bibitem[{Lewis et~al.(2020{\natexlab{b}})Lewis, Perez, Piktus, Petroni,
  Karpukhin, Goyal, K{\"u}ttler, Lewis, Yih, Rockt{\"a}schel
  et~al.}]{lewis2020retrieval}
Patrick Lewis, Ethan Perez, Aleksandara Piktus, Fabio Petroni, Vladimir
  Karpukhin, Naman Goyal, Heinrich K{\"u}ttler, Mike Lewis, Wen-tau Yih, Tim
  Rockt{\"a}schel, et~al. 2020{\natexlab{b}}.
\newblock Retrieval-augmented generation for knowledge-intensive nlp tasks.
\newblock In \emph{Neurips}.

\bibitem[{Lewis et~al.(2020{\natexlab{c}})Lewis, Stenetorp, and
  Riedel}]{lewis2020question}
Patrick Lewis, Pontus Stenetorp, and Sebastian Riedel. 2020{\natexlab{c}}.
\newblock Question and answer test-train overlap in open-domain question
  answering datasets.
\newblock \emph{arXiv preprint arXiv:2008.02637}.

\bibitem[{Li et~al.(2020)Li, Min, Iyer, Mehdad, and Yih}]{li2020efficient}
Belinda~Z Li, Sewon Min, Srinivasan Iyer, Yashar Mehdad, and Wen-tau Yih. 2020.
\newblock Efficient one-pass end-to-end entity linking for questions.
\newblock In \emph{EMNLP}.

\bibitem[{Liang et~al.(2016)Liang, Berant, Le, Forbus, and
  Lao}]{liang2016neural}
Chen Liang, Jonathan Berant, Quoc Le, Kenneth~D Forbus, and Ni~Lao. 2016.
\newblock Neural symbolic machines: Learning semantic parsers on freebase with
  weak supervision.
\newblock \emph{arXiv preprint arXiv:1611.00020}.

\bibitem[{Lin et~al.(2020)Lin, Socher, and Xiong}]{lin2020bridging}
Xi~Victoria Lin, Richard Socher, and Caiming Xiong. 2020.
\newblock Bridging textual and tabular data for cross-domain text-to-sql
  semantic parsing.
\newblock In \emph{Findings of EMNLP}.

\bibitem[{Lin et~al.(2018)Lin, Wang, Zettlemoyer, and Ernst}]{lin2018nl2bash}
Xi~Victoria Lin, Chenglong Wang, Luke Zettlemoyer, and Michael~D Ernst. 2018.
\newblock Nl2bash: A corpus and semantic parser for natural language interface
  to the linux operating system.
\newblock In \emph{LREC}.

\bibitem[{Loula et~al.(2018)Loula, Baroni, and Lake}]{loula2018rearranging}
Joao Loula, Marco Baroni, and Brenden~M Lake. 2018.
\newblock Rearranging the familiar: Testing compositional generalization in
  recurrent networks.
\newblock In \emph{EMNLP Blackbox NLP Workshop}.

\bibitem[{McCloskey and Cohen(1989)}]{mccloskey1989catastrophic}
Michael McCloskey and Neal~J Cohen. 1989.
\newblock Catastrophic interference in connectionist networks: The sequential
  learning problem.
\newblock In \emph{Psychology of learning and motivation}.

\bibitem[{Min et~al.(2013)Min, Grishman, Wan, Wang, and
  Gondek}]{min2013distant}
Bonan Min, Ralph Grishman, Li~Wan, Chang Wang, and David Gondek. 2013.
\newblock Distant supervision for relation extraction with an incomplete
  knowledge base.
\newblock In \emph{NAACL-HLT}.

\bibitem[{Min et~al.(2019)Min, Zhong, Zettlemoyer, and
  Hajishirzi}]{min-etal-2019-multi}
Sewon Min, Victor Zhong, Luke Zettlemoyer, and Hannaneh Hajishirzi. 2019.
\newblock \href {https://doi.org/10.18653/v1/P19-1613} {Multi-hop reading
  comprehension through question decomposition and rescoring}.
\newblock In \emph{Proceedings of the 57th Annual Meeting of the Association
  for Computational Linguistics}, pages 6097--6109, Florence, Italy.
  Association for Computational Linguistics.

\bibitem[{Ngomo(2018)}]{ngomo20189th}
Ngonga Ngomo. 2018.
\newblock 9th challenge on question answering over linked data (qald-9).
\newblock \emph{language}.

\bibitem[{Nickel et~al.(2011)Nickel, Tresp, and Kriegel}]{nickel2011three}
Maximilian Nickel, Volker Tresp, and Hans-Peter Kriegel. 2011.
\newblock A three-way model for collective learning on multi-relational data.
\newblock In \emph{ICML}.

\bibitem[{Pandey et~al.(2018)Pandey, Contractor, Kumar, and
  Joshi}]{pandey-etal-2018-exemplar}
Gaurav Pandey, Danish Contractor, Vineet Kumar, and Sachindra Joshi. 2018.
\newblock Exemplar encoder-decoder for neural conversation generation.
\newblock In \emph{ACL}.

\bibitem[{Pasupat and Liang(2015)}]{pasupat2015compositional}
Panupong Pasupat and Percy Liang. 2015.
\newblock Compositional semantic parsing on semi-structured tables.
\newblock In \emph{ACL}.

\bibitem[{Peng et~al.(2019)Peng, Parikh, Faruqui, Dhingra, and
  Das}]{peng-etal-2019-text}
Hao Peng, Ankur Parikh, Manaal Faruqui, Bhuwan Dhingra, and Dipanjan Das. 2019.
\newblock Text generation with exemplar-based adaptive decoding.
\newblock In \emph{NAACL}.

\bibitem[{Perez et~al.(2020)Perez, Lewis, Yih, Cho, and
  Kiela}]{perez-etal-2020-unsupervised}
Ethan Perez, Patrick Lewis, Wen-tau Yih, Kyunghyun Cho, and Douwe Kiela. 2020.
\newblock \href {https://doi.org/10.18653/v1/2020.emnlp-main.713} {Unsupervised
  question decomposition for question answering}.
\newblock In \emph{Proceedings of the 2020 Conference on Empirical Methods in
  Natural Language Processing (EMNLP)}, pages 8864--8880, Online. Association
  for Computational Linguistics.

\bibitem[{Polu and Sutskever(2020)}]{polu2020generative}
Stanislas Polu and Ilya Sutskever. 2020.
\newblock \href {http://arxiv.org/abs/2009.03393} {Generative language modeling
  for automated theorem proving}.

\bibitem[{Raffel et~al.(2020)Raffel, Shazeer, Roberts, Lee, Narang, Matena,
  Zhou, Li, and Liu}]{t5}
Colin Raffel, Noam Shazeer, Adam Roberts, Katherine Lee, Sharan Narang, Michael
  Matena, Yanqi Zhou, Wei Li, and Peter~J. Liu. 2020.
\newblock Exploring the limits of transfer learning with a unified text-to-text
  transformer.
\newblock \emph{JMLR}.

\bibitem[{Rissland(1983)}]{rissland1983examples}
Edwina~L Rissland. 1983.
\newblock Examples in legal reasoning: Legal hypotheticals.
\newblock In \emph{IJCAI}.

\bibitem[{Ross(1984)}]{ross1984remindings}
Brian~H Ross. 1984.
\newblock Remindings and their effects in learning a cognitive skill.
\newblock \emph{Cognitive psychology}.

\bibitem[{Saxena et~al.(2020)Saxena, Tripathi, and
  Talukdar}]{saxena2020improving}
Apoorv Saxena, Aditay Tripathi, and Partha Talukdar. 2020.
\newblock Improving multi-hop question answering over knowledge graphs using
  knowledge base embeddings.
\newblock In \emph{ACL}.

\bibitem[{Schank(1982)}]{schank1982dynamic}
Roger~C Schank. 1982.
\newblock \emph{Dynamic memory: A theory of reminding and learning in computers
  and people}.
\newblock cambridge university press.

\bibitem[{Schmidt et~al.(1990)Schmidt, Norman, and
  Boshuizen}]{schmidt1990cognitive}
Henk Schmidt, Geoffrey Norman, and Henny Boshuizen. 1990.
\newblock A cognitive perspective on medical expertise: theory and
  implications.
\newblock \emph{Academic medicine}.

\bibitem[{Shaw et~al.(2020)Shaw, Chang, Pasupat, and
  Toutanova}]{shaw2020compositional}
Peter Shaw, Ming-Wei Chang, Panupong Pasupat, and Kristina Toutanova. 2020.
\newblock Compositional generalization and natural language variation: Can a
  semantic parsing approach handle both?
\newblock \emph{arXiv preprint arXiv:2010.12725}.

\bibitem[{Soares et~al.(2019)Soares, FitzGerald, Ling, and
  Kwiatkowski}]{soares2019matching}
Livio~Baldini Soares, Nicholas FitzGerald, Jeffrey Ling, and Tom Kwiatkowski.
  2019.
\newblock Matching the blanks: Distributional similarity for relation learning.
\newblock In \emph{ACL}.

\bibitem[{Socher et~al.(2013)Socher, Chen, Manning, and
  Ng}]{socher2013reasoning}
Richard Socher, Danqi Chen, Christopher~D Manning, and Andrew Ng. 2013.
\newblock Reasoning with neural tensor networks for knowledge base completion.
\newblock In \emph{Neurips}.

\bibitem[{Su et~al.(2016)Su, Sun, Sadler, Srivatsa, G{\"u}r, Yan, and
  Yan}]{su2016generating}
Yu~Su, Huan Sun, Brian Sadler, Mudhakar Srivatsa, Izzeddin G{\"u}r, Zenghui
  Yan, and Xifeng Yan. 2016.
\newblock On generating characteristic-rich question sets for qa evaluation.
\newblock In \emph{EMNLP}.

\bibitem[{Suhr et~al.(2020)Suhr, Chang, Shaw, and Lee}]{suhr2020exploring}
Alane Suhr, Ming-Wei Chang, Peter Shaw, and Kenton Lee. 2020.
\newblock Exploring unexplored generalization challenges for cross-database
  semantic parsing.
\newblock In \emph{ACL}.

\bibitem[{Sun et~al.(2019{\natexlab{a}})Sun, Bedrax-Weiss, and
  Cohen}]{sun2019pullnet}
Haitian Sun, Tania Bedrax-Weiss, and William~W Cohen. 2019{\natexlab{a}}.
\newblock Pullnet: Open domain question answering with iterative retrieval on
  knowledge bases and text.
\newblock In \emph{EMNLP}.

\bibitem[{Sun et~al.(2018)Sun, Dhingra, Zaheer, Mazaitis, Salakhutdinov, and
  Cohen}]{sun2018open}
Haitian Sun, Bhuwan Dhingra, Manzil Zaheer, Kathryn Mazaitis, Ruslan
  Salakhutdinov, and William~W Cohen. 2018.
\newblock Open domain question answering using early fusion of knowledge bases
  and text.
\newblock In \emph{EMNLP}.

\bibitem[{Sun et~al.(2019{\natexlab{b}})Sun, Deng, Nie, and
  Tang}]{sun2019rotate}
Zhiqing Sun, Zhi-Hong Deng, Jian-Yun Nie, and Jian Tang. 2019{\natexlab{b}}.
\newblock Rotate: Knowledge graph embedding by relational rotation in complex
  space.
\newblock In \emph{ICLR}.

\bibitem[{Talmor and Berant(2018)}]{Talmor2018TheWA}
Alon Talmor and Jonathan Berant. 2018.
\newblock The web as a knowledge-base for answering complex questions.
\newblock In \emph{NAACL-HLT}.

\bibitem[{Toutanova and Chen(2015)}]{toutanova-chen-2015-observed}
Kristina Toutanova and Danqi Chen. 2015.
\newblock Observed versus latent features for knowledge base and text
  inference.
\newblock In \emph{Continuous Vector Space Models and their Compositionality
  Workshop}.

\bibitem[{Velickovic et~al.(2018)Velickovic, Cucurull, Casanova, Romero,
  Li{\`o}, and Bengio}]{Velickovic2018GraphAN}
Petar Velickovic, Guillem Cucurull, A.~Casanova, A.~Romero, P.~Li{\`o}, and
  Yoshua Bengio. 2018.
\newblock Graph attention networks.
\newblock In \emph{ICLR}.

\bibitem[{Wang et~al.(2020)Wang, Shin, Liu, Polozov, and Richardson}]{rat-sql}
Bailin Wang, Richard Shin, Xiaodong Liu, Oleksandr Polozov, and Matthew
  Richardson. 2020.
\newblock {RAT-SQL}: Relation-aware schema encoding and linking for
  text-to-{SQL} parsers.
\newblock In \emph{ACL}.

\bibitem[{Weston et~al.(2018)Weston, Dinan, and
  Miller}]{weston-etal-2018-retrieve}
Jason Weston, Emily Dinan, and Alexander Miller. 2018.
\newblock Retrieve and refine: Improved sequence generation models for
  dialogue.
\newblock In \emph{ConvAI Workshop EMNLP}.

\bibitem[{Wiseman and Stratos(2019)}]{wiseman-stratos-2019-label}
Sam Wiseman and Karl Stratos. 2019.
\newblock Label-agnostic sequence labeling by copying nearest neighbors.
\newblock In \emph{ACL}.

\bibitem[{Wolfson et~al.(2020)Wolfson, Geva, Gupta, Goldberg, Gardner, Deutch,
  and Berant}]{wolfson2020break}
Tomer Wolfson, Mor Geva, Ankit Gupta, Yoav Goldberg, Matt Gardner, Daniel
  Deutch, and Jonathan Berant. 2020.
\newblock Break it down: A question understanding benchmark.
\newblock \emph{Transactions of the Association for Computational Linguistics},
  8:183--198.

\bibitem[{Yih et~al.(2016)Yih, Richardson, Meek, Chang, and Suh}]{yih2016value}
Wen-tau Yih, Matthew Richardson, Christopher Meek, Ming-Wei Chang, and Jina
  Suh. 2016.
\newblock The value of semantic parse labeling for knowledge base question
  answering.
\newblock In \emph{ACL}.

\bibitem[{Yu et~al.(2013)Yu, Yao, Su, Li, and Seide}]{yu2013kl}
Dong Yu, Kaisheng Yao, Hang Su, Gang Li, and Frank Seide. 2013.
\newblock Kl-divergence regularized deep neural network adaptation for improved
  large vocabulary speech recognition.
\newblock In \emph{ICASSP}.

\bibitem[{Yu et~al.(2018)Yu, Zhang, Yang, Yasunaga, Wang, Li, Ma, Li, Yao,
  Roman et~al.}]{yu2018spider}
Tao Yu, Rui Zhang, Kai Yang, Michihiro Yasunaga, Dongxu Wang, Zifan Li, James
  Ma, Irene Li, Qingning Yao, Shanelle Roman, et~al. 2018.
\newblock Spider: A large-scale human-labeled dataset for complex and
  cross-domain semantic parsing and text-to-sql task.
\newblock In \emph{EMNLP}.

\bibitem[{Zaheer et~al.(2020)Zaheer, Guruganesh, Dubey, Ainslie, Alberti,
  Ontanon, Pham, Ravula, Wang, Yang et~al.}]{zaheer2020big}
Manzil Zaheer, Guru Guruganesh, Avinava Dubey, Joshua Ainslie, Chris Alberti,
  Santiago Ontanon, Philip Pham, Anirudh Ravula, Qifan Wang, Li~Yang, et~al.
  2020.
\newblock Big bird: Transformers for longer sequences.
\newblock In \emph{Neurips}.

\bibitem[{Zelle and Mooney(1996)}]{zelle1996learning}
John~M Zelle and Raymond~J Mooney. 1996.
\newblock Learning to parse database queries using inductive logic programming.
\newblock In \emph{NCAI}.

\bibitem[{Zhang et~al.(2018)Zhang, Utiyama, Sumita, Neubig, and
  Nakamura}]{zhang2018guiding}
Jingyi Zhang, Masao Utiyama, Eiichro Sumita, Graham Neubig, and Satoshi
  Nakamura. 2018.
\newblock Guiding neural machine translation with retrieved translation pieces.
\newblock In \emph{NAACL}.

\bibitem[{Zhong et~al.(2017{\natexlab{a}})Zhong, Xiong, and
  Socher}]{zhong2017seq2sql}
Victor Zhong, Caiming Xiong, and Richard Socher. 2017{\natexlab{a}}.
\newblock Seq2sql: Generating structured queries from natural language using
  reinforcement learning.
\newblock \emph{arXiv preprint arXiv:1709.00103}.

\bibitem[{Zhong et~al.(2017{\natexlab{b}})Zhong, Xiong, and
  Socher}]{zhongSeq2SQL2017}
Victor Zhong, Caiming Xiong, and Richard Socher. 2017{\natexlab{b}}.
\newblock Seq2sql: Generating structured queries from natural language using
  reinforcement learning.
\newblock \emph{CoRR}, abs/1709.00103.

\end{thebibliography}
\bibliographystyle{acl_natbib}

\appendix

\clearpage

\section{EMNLP Reproducibility Checklist}
\label{sec:appendix_data_hyper_param}
\subsection{Data}
\label{sub:appendix_data}
WebQSP contains 4737 NL questions belonging to 56 domains covering 661 unique relations. Most questions need up to 2 hops of reasoning, where each hop is a KB edge. \cwq (CWQ) is generated by extending the WebQSP dataset with the goal of making it a more complex multi-hop dataset. There are four types of questions: composition (45\%), conjunction (45\%), comparative (5\%), and superlative (5\%). Answering these questions requires up to 4 hops of reasoning in the KB, making the dataset challenging. Compositional Freebase Questions (CFQ) is a recently proposed benchmark explicitly developed for measuring compositional generalization. For all the datasets above, the logical form (LF) for each NL question is a \textsc{Sparql} query that can be executed against the Freebase KB to obtain the answer entity.

\begin{table}[]
    \centering
    \begin{tabular}{l c c c}\toprule
         Dataset & Train & Valid & Test \\\midrule
         WebQSP & 2,798 & 300 & 1,639\\
         CWQ & 27,639 & 3,519 & 3,531 \\
         CFQ & 95,743 & 11,968 & 11,968 \\\bottomrule
         
    \end{tabular}
    \caption{Dataset statistics}
    \label{tab:data_stats}
\end{table}

\subsection{Hyperparameters}
\label{sub:appendix_hyperparam}

The WebQSP dataset does not contain a validation split, so we choose 300 training instances to form the validation set. We use grid-search (unless explicitly mentioned) to set the hyperparameters listed below.\\
\noindent\textbf{Case Retriever}: We initialize our retriever with the pre-trained \roberta-base weights. We set the initial learning rate to $5 \times 10^{-5}$ and decay it linearly throughout training. We evaluate the retriever based on the percentage of gold LF relations in the LFs of the top-k retrieved cases (recall@k). We train for 10 epochs and use the best checkpoint based on recall@20 on the validation set. We set train and validation batch sizes to 32.

For $p_{mask}$, we try values from [0, 0.2, 0.4, 0.5, 0.7, 0.9, 1]. When training the retriever, we found $p_{\textrm{mask}}=0.2$ works best for \cwq and $p_{\textrm{mask}}=0.5$ for the remaining datasets.

\textbf{Seq2Seq Generator}: We use a \bigb generator network with 6 encoding and 6 decoding sparse-attention layers, which we initialize with pre-trained \textsc{Bart}-base weights. We set the initial learning rate to $5\times 10^{-5}$ and decay it linearly throughout training. Accuracy after the execution of generated programs on the validation set is used to select the optimal setting and model checkpoint.

For $\lambda_T$, we perform random search in range [0, 1]. We finally use $\lambda_T$=1.0 for all datasets. For $k$ (number of cases), we search over the values [1, 3, 5, 7, 10, 20]. For all datasets, we use $k$=20 cases and decode with a beam size of 5 for decoding. The WebQSP model was trained for 15K gradient steps and all other models were trained for 40K gradient steps.

\textbf{Computing infrastructure}: We perform our experiments on a GPU cluster managed by SLURM. The case retriever was trained and evaluated on NVIDIA GeForce RTX 2080 Ti GPU. The models for the Reuse step were trained and evaluated on NVIDIA GeForce RTX 8000 GPUs. Revise runs on NVIDIA GeForce RTX 2080 Ti GPU when using \textsc{Roberta} for alignment and runs only on CPU when using \textsc{TransE}. We report validation set scores in Table~\ref{tab:dev_results}.

\begin{table}[]
    \centering
    \begin{tabular}{l c}\toprule
         Dataset & Validation Acc \\\midrule
         WebQSP & 71.5\\
         CWQ & 82.8 \\
         CFQ & 69.9 \\\bottomrule
    \end{tabular}
    \caption{Validation set accuracy of models corresponding to the results reported in the paper}
    \label{tab:dev_results}
\end{table}

\section{Further Experiments and Analysis}
\label{sec:appendix_further_experiments}

\subsection{Performance of Retriever}
\label{sub:appendix_retriever}
We compare the performance of our trained retriever with a \roberta-base model. We found that \roberta model even without any fine-tuning performs well at retrieval. However, fine-tuning \roberta with our distant supervision objective improved the overall recall, e.g., from 86.6\% to 90.4\% on \webqsp and from 94.8\% to 98.4\% on CFQ.

\subsection{Performance on Unseen Entities}
In Table~\ref{tab:hitl_results_appendix} we showed \alg is effective for unseen relations. But what about unseen entities in the test set?. On analysis we found that in WebQSP, \alg can copy unseen entities correctly \textbf{86.8}\% (539/621) from the question. This is +1.9\% improvement from baseline transformer model which is able to copy correctly 84.9\% (527/621) of the time. Note that unseen entities can be copied from the input NL query and we do not need additional cases to be injected to KNN index.

\subsection{Analysis of the Revise Step}

In the revise step, we attempt to fix programs predicted by our reuse step that did not execute on the knowledge base. The predicted program can be syntactically incorrect or enforce conditions that lead to an unsatisfiable query. In our work, we focus on predicted programs that can be fixed by aligning clauses to relations in the local neighborhood of query entities. We give examples of successful alignments Table~\ref{tab:revise_eye_candies_pos_appendix} as well as failed attempts at alignment Table~\ref{tab:revise_eye_candies_neg_appendix}.

\section{Details on Held-Out Experiments}
\label{sub:held_out_appendix}

In this section, we include more details about our held-out experiment described in section~\ref{subsub:unseen_rel}. The goal of this experiment is to show that our approach can \emph{generalize} to unseen relations without requiring \emph{any further training} of the model.
This is a relevant setting to explore, because real-world knowledge bases are often updated with new kinds of relations, and we would like KBQA systems that adapt to handle new information with minimal effort.

We explicitly hold-out all questions containing a particular relation from the datasets. Table~\ref{tab:held_out_stats} shows the relation type and the number of questions that are removed as a result of removing the relation.

\begin{table}[bht]
    \centering
    \small
    \begin{tabular}{c c c c} \toprule
         Dataset & Relation name & Train & Test\\\midrule
         WebQSP & people.person.education & 136 & 86 \\\bottomrule
    \end{tabular}
    \caption{Relation type and the corresponding number of NL queries that are held-out.}
    \label{tab:held_out_stats}
\end{table}

\section{Details on Automated Case Collection and Human-in-the-Loop Experiments}
\label{app:hitl}
\begin{table*}
    \footnotesize
    \centering
    \begin{tabular}{c c c c c c} \toprule
    & & & \multicolumn{2}{ c }{Cases Added via}\\ \cline{4-5}
         Dataset & \# missing relations  & \# questions & \hitl & SimpleQuestions & Avg. \# cases per relation \\\midrule
         WebQSP & 94 & 79 & 72 & 292 & 3.87 \\\bottomrule
    \end{tabular}
    \caption{Number of questions in the evaluation set that needs a relation which is not seen in the training set. Note that, there can be multiple relations in a question that might not be seen during training. The last two columns show the number of cases added both via human-in-the-loop (\hitl) annotation and automatically from SimpleQuestions dataset. }
    \label{tab:hilt_stats_appendix}
\end{table*}

\begin{table*}
    \centering
    \small
    \setlength{\tabcolsep}{3pt}
    \begin{tabular}{@{}l l c@{}}\toprule
        NL Query & \spql & Source\\\midrule
        What is the Mexican Peso called? & select ?x where \{ m.012ts8 finance.currency.currency\_code ?x .\} & Manual\\
        Who invented the telephone? & select ?x where \{ m.0g\_3r  base.argumentmaps.original\_idea.innovator ?x  .\} & Manual \\
        what area is wrvr broadcated in? & select ?x where \{ m.025z9rx broadcast.broadcast.area\_served ?x .\} & SQ\\
        Where are Siamese cats originally from? & select ?x where \{ m.012ts8 biology.animal\_breed.place\_of\_origin ?x .\} & Manual\\ \bottomrule   
    \end{tabular}
    \caption{Examples of few added questions and their corresponding \spql queries. Notice that the \spql queries are very simple to create once we know the name of the missing relation. The source column indicate whether the question was manually created or automatically added from Simple Questions (SQ) dataset.}
    \label{tab:hitl_appendix}
\end{table*}

While conducting analysis, we also noticed that WebQSP has queries in the test set for which the required relations are never present in the training set. This gives us an opportunity to conduct real human-in-the-loop experiments to demonstrate the advantage of our model. To add more cases, we resort to a mix of automated data collection and human-in-the-loop strategy. For each of the missing relation, we first try to find NL queries present in the SimpleQuestions \cite{bordes2015large}. SimpleQuestions (SQ) is a large dataset containing more than 100K NL questions that are `simple' in nature --- i.e. each NL query maps to a single relation (fact) in the Freebase KB.  For each missing relation type, we try to find questions in the SQ dataset that can be mapped to the missing relation. However, even SQ has missing coverage in which case, we manually generate a question and its corresponding \spql query by reading the description of the relation. Table~\ref{tab:hilt_stats_appendix} shows the number of questions in the evaluation set which at least has a relation never seen during training and also the number of cases that has been added. For example, we\footnote{The H-I-T-L case addition was done by 2 graduate students in the lab. } were able to collect 292 questions from SQ and we manually created 72 questions for WebQSP. Overall, we add 3.87 new cases per query relation for WebQSP.

Table~\ref{tab:hitl_appendix} shows some example of cases added manually or from SQ. We look up entity ids for entities from the FACC1 alias table (\S\ref{sub:entity_linking}). Also note, that since we only add questions which are simple in nature, the corresponding \spql query can be easily constructed from the missing relation type and the entity id.

\begin{figure}
    \centering
    \vspace{-3mm}
    \includegraphics[width=\columnwidth]{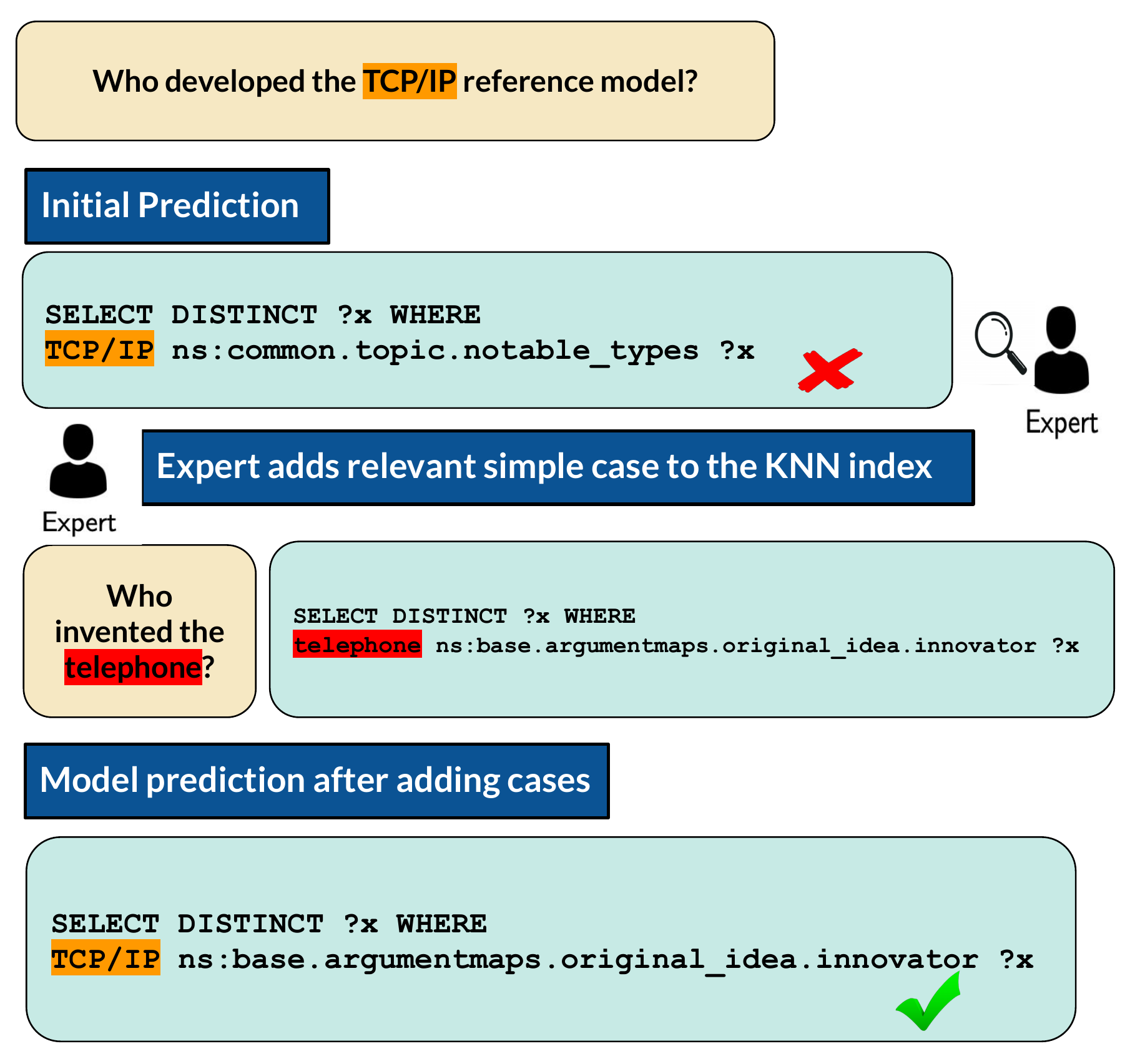}
    \caption{\small An example query where our approach correctly utilizes added H-I-L-T cases}
    \vspace{-5mm}
    \label{fig:hitl_ec1}
\end{figure}
\begin{figure}
    \centering
    \vspace{-3mm}
    \includegraphics[width=\columnwidth]{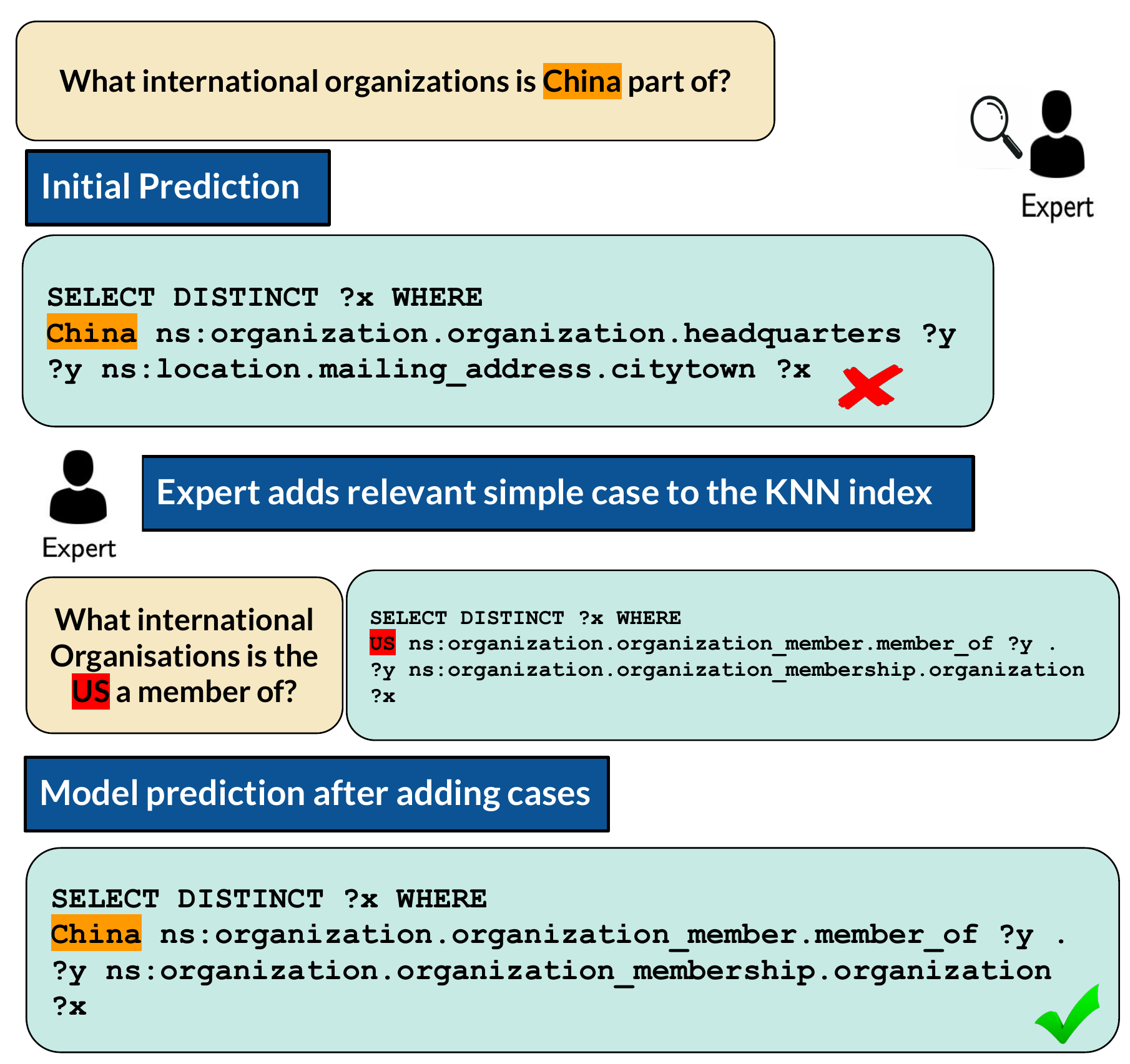}
    \caption{\small An example query where our approach correctly utilizes added H-I-L-T cases}
    \vspace{-5mm}
    \label{fig:hitl_ec2}
\end{figure}

\textbf{Importance of this result}: Through this experiment, we demonstrate two important properties of our model --- \emph{interpretability} and \emph{controllability}. Database schemas keep changing and new tables keep getting added to a corporate database. When our QA system gets a query wrong, by looking at the retrieved K-nearest neighbors, users can determine (interpretability) that the required relation is not present in the training set. By adding few cases for the new relations, they can query the DB for similar questions, without needing to train the QA system (controllability). Current black-box NLP models are not capable of doing such point-fixes and our experiment is an initial attempt towards building such systems. 

\section{Further Related Work}
\label{sec:further_related_work}
\paragraph{\textbf{KNN approach in other NLP applications (continued)}:}
\citet{wiseman-stratos-2019-label} achieved accurate sequence labeling by explicitly and only copying labels from retrieved neighbors. NN models have been used in a numbe NLP applications such as POS tagging \cite{daelemans1996mbt}. Another recent line of work use training examples at test time to improve language generation \cite{weston-etal-2018-retrieve,pandey-etal-2018-exemplar,cao-etal-2018-retrieve,peng-etal-2019-text}. \citet{hua2020retrieve} recently proposed a meta-learning approach which utilizes cases retrieved w.r.t. the similarity of the input. However, their main goal is to learn a better parametric model (retriever and generator) from neighboring cases rather than composing and fixing cases to generate answers at test time.

\paragraph{\textbf{Question Decomposition}}
One strategy to answer a complex question is to first break it down into simpler subquestions, each of which can be viewed as a natural language program describing how to answer the question.
This approach has been shown to be effective as far back as IBM Watson~\cite{ferrucci2010building} to more recent systems for answering questions about text~\cite{das2019multi,min-etal-2019-multi,perez-etal-2020-unsupervised,wolfson2020break} or knowledge bases~\cite{Talmor2018TheWA}.
These prior studies do not leverage case-based reasoning when generating decompositions and thus may also benefit from similar techniques as proposed in our work.

\begin{table*}
    \centering
    \small
    \renewcommand{\arraystretch}{1.5}
    \begin{tabular}{l l}
    \toprule
     & WebQSP\\
    \hline
    Question: & when did kaley cuoco m.03kxp7 join charmed m.01f3p\_ ?\\
    Predicted \spql: & \begin{minipage}[t]{1.5\columnwidth}\textsf{SELECT DISTINCT ?x WHERE \{\\\quad ns:m.03kxp7 ns:tv.tv\_character.appeared\_in\_tv\_program ?y .\\\quad ?y ns:tv.regular\_tv\_appearance.from ?x .\\\quad ?y ns:tv.regular\_tv\_appearance.series ns:m.01f3p\_ .\\ \}}\end{minipage}\\
    Ground-truth \spql: & \begin{minipage}[t]{1.5\columnwidth}\textsf{SELECT DISTINCT ?x WHERE \{\\\quad ns:m.03kxp7 ns:tv.tv\_actor.starring\_roles ?y .\\\quad ?y ns:tv.regular\_tv\_appearance.from ?x .\\\quad ?y ns:tv.regular\_tv\_appearance.s
eries ns:m.01f3p\_ .\\ \}}\end{minipage}\\
    Revised \spql: & \begin{minipage}[t]{1.5\columnwidth}\textsf{SELECT DISTINCT ?x WHERE \{\\\quad ns:m.03kxp7 \textbf{ns:tv.tv\_actor.starring\_roles} ?y .\\\quad ?y ns:tv.regular\_tv\_appearance.from ?x .\\\quad ?y ns:tv.regular\_tv\_appearance.s
eries ns:m.01f3p\_ .\\ \}}\end{minipage}\\
    \midrule
     & CWQ\\
    \hline
    Question: & \begin{minipage}[t]{1.5\columnwidth}What text in the religion which include Zhang Jue m.02gjv7 as a key figure is considered to be sacred m.02vt2rp ? \end{minipage}\\
    Predicted \spql: & \begin{minipage}[t]{1.5\columnwidth}\textsf{SELECT DISTINCT ?x WHERE \{\\\quad ?c ns:religion.religion.deities ns:m.02gjv7 .\\\quad ?c ns:religion.religion.texts ?x .\\\quad \ldots \textit{benign filters} \ldots \}\\\quad }\end{minipage}\\
    Ground-truth \spql: & \begin{minipage}[t]{1.5\columnwidth}\textsf{SELECT DISTINCT ?x WHERE \{\\\quad ?c ns:religion.religion.notable\_figures ns:m.02gjv7 .\\\qquad ?c ns:religion.religion.texts ?x .\}\\\quad }\end{minipage}\\
    Revised \spql: & \begin{minipage}[t]{1.5\columnwidth}\textsf{SELECT DISTINCT ?x WHERE \{\\\quad ?c \textbf{ns:religion.religion.notable\_figures} ns:m.02gjv7 .\\\quad ?c ns:religion.religion.texts ?x .\\\quad \ldots \textit{benign filters} \ldots \}\\\quad }\end{minipage}\\
    \midrule
    Question: & \begin{minipage}[t]{1.5\columnwidth} What is the mascot of the educational institution that has a sports team named the North Dakota State Bison m.0c5s26 ? \end{minipage}\\
    Predicted \spql: & \begin{minipage}[t]{1.5\columnwidth}\textsf{SELECT DISTINCT ?x WHERE \{\\\quad ?c ns:education.educational\_institution.sports\_teams ns:m.0c5s26 .\\\quad ?c ns:education.educational\_institution.mascot ?x .\\\quad\} }\end{minipage}\\
    Ground-truth \spql: & \begin{minipage}[t]{1.5\columnwidth}\textsf{SELECT DISTINCT ?x WHERE \{\\\quad ?c ns:education.educational\_institution.sports\_teams ns:m.0c41\_v .\\\quad ?c ns:education.educational\_institution.mascot ?x .\\\quad\} }\end{minipage}\\
    Revised \spql: & \begin{minipage}[t]{1.5\columnwidth}\textsf{SELECT DISTINCT ?x WHERE \{\\\quad ?c \textbf{ns:education.educational\_institution.athletics\_brand} ns:m.0c5s26 .\\\quad ?c ns:education.educational\_institution.mascot ?x .\\\quad\} }\end{minipage}\\
    Comments: & \begin{minipage}[t]{1.5\columnwidth} The entity linker has tagged the bison as a university symbol (m.0c5s26) rather than the Bison football team (m.0c41\_v). Alignment helps the model recover from this by picking the relation that connects the tagged entity to the university. \end{minipage}\\
    \bottomrule
    \end{tabular}
    \caption{Examples of successful alignment with TransE from the Revise stage.}
    \label{tab:revise_eye_candies_pos_appendix}
\end{table*}

\begin{table*}
    \centering
    \small
    \renewcommand{\arraystretch}{1.5}
    \begin{tabular}{l@{}l}
    \toprule
     & WebQSP\\
    \hline
    Question: & \begin{minipage}[t]{1.5\columnwidth} who is gimli m.0h34n 's father m.02pn7 in the hobbit m.0n4ck66 \end{minipage}\\
    Predicted \spql: & \begin{minipage}[t]{1.5\columnwidth}\textsf{SELECT DISTINCT ?x WHERE \{\\\quad ns:m.0h34n ns:people.person.parents ?x .\\\quad ?x ns:people.person.gender ns:m.02pn7 .\\\quad ?x ns:people.person.parents ?sk0 .\\ \} ORDER BY xsd:datetime(?sk0) LIMIT 1 }\end{minipage}\\
    Ground-truth \spql: & \begin{minipage}[t]{1.5\columnwidth}\textsf{SELECT DISTINCT ?x WHERE \{\\\quad ns:m.0h34n ns:fictional\_universe.fictional\_character.parents ?x .\\\quad ?x ns:fictional\_universe.fictional\_character.gender ns:m.05zppz .\\ \}} \end{minipage}\\
    Predicted \spql: & \begin{minipage}[t]{1.5\columnwidth}\textsf{SELECT DISTINCT ?x WHERE \{\\\quad ns:m.0h34n \textbf{ns:fictional\_universe.fictional\_character.parents} ?x .\\\quad ?x ns:people.person.gender ns:m.02pn7 .\\\quad ?x ns:people.person.parents ?sk0 .\\ \} ORDER BY xsd:datetime(?sk0) LIMIT 1 }\end{minipage}\\
    Comments: & \begin{minipage}[t]{1.5\columnwidth} In this example the prediction has an incorrect structure, so aligning an edge does not change the outcome. \end{minipage}\\
    \midrule
     & CWQ\\
    \hline
    Question: & \begin{minipage}[t]{1.5\columnwidth} What political leader runs the country where the Panama m.05qx1 nian Balboa m.0200cp is used? \end{minipage}\\
    Predicted \spql: & \begin{minipage}[t]{1.5\columnwidth}\textsf{SELECT DISTINCT ?x WHERE \{\\\quad ?c ns:location.country.currency\_formerly\_used ns:m.0200cp .\\\quad ?c ns:government.governmental\_jurisdiction.governing\_officials ?y .\\\quad
    ?y ns:government.government\_position\_held.office\_holder ?x .\\\quad \ldots \textit{benign filters} \ldots \}\\\quad }\end{minipage}\\
    Ground-truth \spql: & \begin{minipage}[t]{1.5\columnwidth}\textsf{SELECT DISTINCT ?x WHERE \{\\\quad ?c ns:location.country.currency\_used ns:m.0200cp .\\\qquad ?c ns:government.governmental\_jurisdiction.governing\_officials ?y .\\\quad ?y ns:government.government\_position\_held.office\_holder ?x .\\\quad \textbf{?y ns:government.government\_position\_held.office\_position\_or\_title ns:m.0m57hp6} .\\\quad \ldots \textit{benign filters} \ldots \}\\\quad }\end{minipage}\\
    Revised \spql: & \begin{minipage}[t]{1.5\columnwidth}\textsf{SELECT DISTINCT ?x WHERE \{\\\quad ?c \textbf{ns:location.country.currency\_used} ns:m.0200cp .\\\quad ?c ns:government.governmental\_jurisdiction.governing\_officials ?y .\\\quad
    ?y ns:government.government\_position\_held.office\_holder ?x .\\\quad \ldots \textit{benign filters} \ldots \}\\\quad }\end{minipage}\\
    Target Answers: & \{m.06zmv9x\} \\
    Revised Answers: & \{m.02y8\_r, m.06zmv9\}\\
    Comments: & \begin{minipage}[t]{1.5\columnwidth} The original prediction has missing clauses so alignment produces more answers than target program\end{minipage}\\
    \bottomrule
    \end{tabular}
    \caption{Examples of failed alignment with TransE from the Revise stage.}
    \label{tab:revise_eye_candies_neg_appendix}
\end{table*}



\end{document}